\documentclass{article} % For LaTeX2e
\usepackage[final]{colm2026/colm2026_conference}

\usepackage{microtype}
\usepackage{hyperref}
\usepackage{url}
\usepackage{booktabs}
\usepackage{listings}
\usepackage{wrapfig}

\usepackage{amssymb}
\usepackage{amsmath}
\usepackage{enumitem}
\usepackage{multicol}
\usepackage{multirow}
\usepackage{graphicx}
\usepackage[table]{xcolor}
\usepackage{colortbl}
\usepackage{subcaption}

\definecolor{customblue}{RGB}{173, 216, 230}

\newif\ifshowchanges
\showchangesfalse % black, turn off edit highlights

\ifshowchanges
    \definecolor{edits}{named}{blue}
\else
    \definecolor{edits}{named}{black}
\fi

\usepackage{lineno}

\definecolor{darkblue}{rgb}{0, 0, 0.5}
\definecolor{customblue}{RGB}{197,235,255}
\hypersetup{colorlinks=true, citecolor=darkblue, linkcolor=darkblue, urlcolor=darkblue}

\title{A Self-Pruning Transformer: Extreme KV-Cache Compression with Universal Attention}

\author{%
    Davis Wertheimer$^{1,}$\thanks{Equal contribution.}\ \ , Haochen Shen$^{2,*}$, Ahan Gupta$^{2,*}$, Derrick Liu$^1$,\\
    \textbf{Yu Chin Fabian Lim$^1$, Mudhakar Srivatsa$^1$, Raghu K. Ganti$^1$,}\\
    \textbf{Minjia Zhang$^2$, Naigang Wang$^1$}\\
    $^1$IBM Research, USA\ $^2$SSAIL Lab, University of Illinois Urbana-Champaign\\
    \texttt{davis.wertheimer@ibm.com} \quad \texttt{\{hshen14,ag82\}@illinois.edu}
}

\begin{document}

\ifcolmsubmission
\linenumbers
\fi

\maketitle

\begin{abstract}
The large KV-cache size of modern LLMs creates a barrier to efficient deployment. Recent work has explored replacing attention layers' RoPE positional embeddings with alternative decay-based mechanisms, which can then be used to prune KV-cache during inference. However, these decay functions have limited expressivity, and in practice devolve into sliding-window-like eviction patterns. In this work, we propose a unifying framework for complementary and novel decay mechanisms, capturing complex key statistics and interactions while preserving expressive RoPE embeddings and Softmax attention. The resulting \emph{Universal Attention} is a highly expressive and end-to-end trainable architecture, whose composite decay mechanism acts as a natural, \textit{adaptive} pruning criterion, removing tokens that contribute least to attention computation. Experimentally, Universal Attention achieves state-of-the-art $10\times$ compression on natural language and synthetic task data, while \textit{improving} downstream performance compared to both state-of-the-art baselines and unpruned oracles. It further demonstrates superior long-context generalization with unprecedented $25\times$ compression at length 16k.

\end{abstract}

\section{Introduction}

% \ahan{As the introduction reads currently, the challenges of what we're doing aren't coming out. As a result, the way its described, it seems like what we're doing is "trivial". TODO: fix.}

% \davis{Reworked the last paragraph to push harder our achievements / contributions}

Large Language Models (LLMs) power many of the applications we use today \citep{gpt5, claude,gemini,grok}. However, they predominantly employ softmax-attention layers~\citep{Vaswani+2017}, which compute pairwise dot-products between \textit{all} tokens in the input sequence. This entails preserving \textit{every} token, for \textit{each} layer, as the ``KV-cache'' during inference time, consuming gigabytes (if not terabytes) of 
% high-bandwidth-memory (HBM) -- a scarce resource on modern hardware. 
device memory, often several times larger than the model itself. 
This acts as a significant obstacle to deploying LLMs efficiently at scale.
% As attention patterns are mostly sparse~\citep{TODO}, a huge portion of the full dot-product lookup operation is wasted computation -- assuming the cache fits in device memory at all.
% This poses two challenges to LLM inference at scale: the KV-cache may not fit in HBM, requiring offloading to other devices, and streaming the cache to and from HBM can bottleneck computation.
% Consequently, this imposes a challenge to foundation model inference at scale, which must effectively solve how to load KV-caches and model state onto HBM during inference.

To effectively reduce KV-cache sizes, recent research has proposed the addition of a ``forget-gate'' to the softmax self-attention formulation, enabling unimportant key/value tokens from the past to be deleted. For example, forgetting transformers~\citep{forgetting_transformer_1, forgetting_transformer_2} (FoX) introduces and applies a scalar data-dependent decay term for each new token entering the cache, accumulating over time as an additive bias to the attention matrix.
% , effectively zeroing certain query-key products after softmax normalization. 
This effectively zeros out certain query-key products after softmax normalization, allowing the corresponding key/value pair to be safely evicted. % from the KV-cache.
% These decay terms are uniquely structured, ensuring that keys which introduce low attention-matrix scores at time-step \texttt{t} can be evicted from the KV-cache alongside their corresponding values. 
% However, existing techniques construct inexpressive forget-gates,
However, these forget-gates are relatively inexpressive,
% that are cheap to compute, 
yielding simple patterns that devolve to soft sliding-windows,
% -like eviction patterns, 
% that delete only a few key/value tokens prior to model quality suffering.
which can only evict limited key/value entries before downstream model performance degrades. %is negatively impacted.

%% Previous paragraph of the above paragraph. %%
%To effectively reduce KV-cache sizes, recent research has proposed the addition of a "forget-gate" to the softmax self-attention formulation, enabling unimportant key/value tokens from the past to be deleted. For example, forgetting transformers~\citep{forgetting_transformer_1, forgetting_transformer_2} (FOX) introduces a scalar data-dependent decay term for each token as an additive bias to the attention-matrix. In doing so, certain query-key products are zeroed after softmax normalization. Moreover, the decay-terms are structured such that if a particular query-key product is zeroed at time step \texttt{t} for a particular key, \texttt{k}, then all future query-key products of \texttt{k} are zeroed, permitting the deletion of \texttt{k} (and its corresponding value-token) from the KV-cache. However, existing techniques construct inexpressive "forget-gates" that are cheap to compute, yielding simple patterns that delete only a few key/value tokens.

In this work, we introduce \emph{Universal Attention} (UA), a unifying framework that combines first- and second-order statistics of stored keys to produce uniquely expressive forget gates.
Some components of our framework align with and generalize decay mechanisms from popular Linear Attention models, such as DeltaNet~\citep{delta_rule,gated_delta_net} \& Mamba~\citep{mamba, mamba2}. 
But while these models rely on a constant-sized \texttt{d$\times$d} state that stores at most \texttt{d} linearly independent key/value associations, we combine these and novel mechanisms into a single pruning technique for the full KV-cache of SoftMax attention, while preserving RoPE embeddings~\citep{rope}.
We additionally implement hardware-aware custom UA kernels for high-throughput training.
The resulting UA architecture is efficient and expressive, achieving over 10$\times$ KV-cache compression on natural language and synthetic task data, for a 1B transformer model trained on 50B tokens.
This memory overhead is lower than state-of-the-art pruning methods and even $O(1)$ linear attention architectures, despite \textit{improved} downstream performance, even compared to unpruned oracle baselines.
% Impact on downstream model quality is minimal, and, interestingly, sometimes even positive, with reasoning and information retrieval scores beating prior state-of-the-art pruning methods, all with lower memory overhead than even O(1) linear attention architectures. 
Cache pruning in UA is also adaptive and input-dependent, generalizing to sequence lengths not observed during training, with unprecedented $25\times$ compression in long contexts of 16k.

\section{Related Work}

Shrinking the KV cache is a topical problem with several orthogonal approaches. These include dedicated architectures with sublinear cache growth at inference time, techniques to reduce KV-cache access without reducing the overall size, post-training pruning techniques for vanilla architectures, and combined architectures that use trained forget gates to decay KV cache entries. UA sits at the intersection of these approaches. 

% \davis{open with a summary line on the multiple orthogonal avenues of exploration wrt memory savings, with UA sitting at the intersection. TODO: come back and take care of this myself.}

\textbf{Sub-quadratic architectures:} Linear attention models collapse the sequence length into a constant, \texttt{d$_{state}\times$d$_{state}$}-sized, recurrent hidden state~\cite{linear-attn}. However, when input sequence lengths exceed \texttt{d$_{state}$}, retrieval from the hidden state becomes noisy. To mitigate this issue, models in this family impose a recency bias by decaying past key/value associations~\citep{mamba,mamba2,gated_delta_net}. 
% Nevertheless, despite introducing expressive decay-mechanisms, 
While still highly expressive,
these models experience considerable drops in quality for complex long-context tasks compared to Softmax self-attention~\citep{repeat-after-me,decimamba}. 

In addition, despite their O(1) memory in the limit, such models can fail to reduce overhead in practice: for typical model configurations, 
% hidden state size for a typical linear attention layer is $2dh$, where $d$ is model width, $h$ is head dimension, typically 128, and 2 is a built-in latent scaling factor. The corresponding attention layer with Group-Query Attention~(GQA) has overhead $2Ld/k$, where $L$ is sequence length, $k$ is the GQA group size (typically 8), and 2 comes from the fact that both keys and values are being stored. 
% Plugging in these typical values and solving for $L$ shows that 
linear attention consumes \textit{greater} memory for sequences shorter than 1k tokens - a sizable portion of the usual training length (see Appendix~\ref{app:kv_cache_comp}).
% TODO: copy above into its own appendix discussion
Fixes aimed at improving model capacity and performance only exacerbate the issue: hybrid models interleaving full attention with linear attention still have the same break-even point, but become O(L) in the limit, reducing best-case memory savings to 8-10$\times$ for typical settings where $1/8$ to $1/10$ of layers are switched to Softmax~\citep{nemotron-h,jamba,jamba1-5}. 
Log-Linear Attention~\citep{log_linear_attention} maintains multiple state buffers, pushing out the break-even point even further.
UA instead applies an expressive decay mechanism to Softmax self-attention directly, avoiding the memory lower-bound imposed by a constant-sized recurrent state. 

\textbf{Sparse attention:} \citet{big_bird, sparse_transformer, native_sparse_attention} compute only a subset of the attention matrix. These techniques range from computing static patterns (such as sliding-windows around the diagonal) to dynamic patterns 
% that select important query-key dot-products through 
based on 
custom data-dependent heuristics. Simple patterns like sliding-window-attention reduce KV-cache sizes but degrade long-context capability unless interleaved with 
more expressive layers~\citep{mistral,longformer,gemma3},
% full SoftMax-attention \textbf{cite} 
{\color{edits}while content-dependent dynamic sparsity provides a more expressive alternative~\citep{dam, dma}}. 
% On the other hand, dynamically sparse 
These dynamic
sparse attention patterns are either implemented by modifying full-attention masks to zero out unnecessary query-key dot-products, or by selectively reading queries and keys from memory to compute a partial attention matrix. Neither implementation reduces KV-cache sizes.
% physically deletes key/value tokens from memory, retaining the full KV-cache. 
% Comparatively, UA provides over 10$\times$ compression without compromising model quality.

\textbf{KV-cache eviction:} Methods such as H$_2$O and SnapKV~\citep{h2o, snapkv} use custom heuristics during the prefill \& decode stages of LLM inference to determine which key/value tokens to evict from the KV-cache. {\color{edits} More recent methods refine the eviction criterion under the same fixed-budget setting: CAOTE~\citep{caote} combines attention scores with value vectors to minimize the attention-output error induced by eviction, while KeyDiff~\citep{keydiff} avoids attention scores and instead retains geometrically distinctive keys based on key-space similarity.} These strategies compress KV-caches to a preset target (e.g., evicting 90$\%$ of tokens), while UA's adaptive pruning dynamically determines which key/value tokens are dispensable. 
As a result, UA compresses much further at long sequence length, in addition to improving performance more generally.

\textbf{Data-dependent forget gates:}
% Similar combinatorial idea, but does not discuss pruning or memory savings. We generalize this approach and analyze downstream behavior more extensively. 
% have been applied to a variety of models such as LSTMs~\citep{}%, feed-forward-networks~\citep{}, 
% and transformers~\citep{Vaswani+2017}. 
A range of gate formulations have been added to a variety of models, such as LSTMs~\citep{LSTM} and transformers~\citep{Vaswani+2017}, for a variety of purposes. 
But from a cache compression perspective, many approaches devolve into soft-sliding-window formulations~\citep{forgetting_transformer_1,forgetting_transformer_2,LSTM}. More expressive mechanisms such as the delta rule~\citep{delta_rule,gated_delta_net,PaTH_attention} preclude the use of well-established elements of attention, such as SoftMax or RoPE embeddings~\citep{rope}. UA combines multiple forget gates into an additive bias to the standard attention matrix, allowing us to leverage rich positional encoding via RoPE, targeted retrieval via SoftMax, and more sophisticated decay patterns than sliding-window (see Section~\ref{sec:empirical}), all at the same time.

\section{Background}
% \haochen{This entire section probably should go to appendix. The introduced linear attention mechanisms seem weakly linked to our method.}

% \davis{Keep the initial section here to establish terminology. Split section 4 into two subsections: decay mechanisms, and composition. Merge the Linear Attention section(s) into the corresponding mechanisms in section 4.1. That way we can more closely link the original DeltaNet formula, and our alteration. This will also help reduce signposting in sec4. We may also be able to drop the Mamba section/equations entirely in favor of FoX, and simply establish the Mamba link through citations.}

% \davis{TODO: I'll take care of merging the deltanet and similarity decay sections, I know what I want to say there.}

% \ahan{Restructured Section 4.0. Left the deltanet decay-mechnaism unchanged (since you have additional things you'd like to say).}

This section introduces the notation and necessary background relevant to UA's design.

\textbf{Softmax multi-head self-attention:} Standard softmax multi-head self-attention takes an input sequence of tokens $(\mathbf{q}_i)_{i=1}^L,  (\mathbf{k}_i)_{i=1}^L, (\mathbf{v}_i)_{i=1}^L\in \mathbb{R}^{n\times d}$, where $L$, $n$ \& $d$ are the sequence length, number of heads, and head dimension, respectively and computes:  
% \begin{equation}
%     \underbrace{\text{Softmax}(\mathbf{Q}\mathbf{K}^\top+ \mathbf{M})}_{\mathbb{R}^{n\times L \times L}}\mathbf{V},
% \end{equation}
\begin{equation}
    \text{Softmax}(\mathbf{Q}\mathbf{K}^\top+ \mathbf{M})\mathbf{V},
\end{equation}
where $\mathbf{M}\in\mathbb{R}^{L\times L}$ is the causal attention mask broadcasted across heads, $\mathbf{Q}, \mathbf{K}, \mathbf{V} \in \mathbb{R}^{n\times L \times d}$ are tensors formed by stacking $\mathbf{q}_i$ , $\mathbf{k}_i$ \& $\mathbf{v}_i$, respectively. During inference, new tokens need to attend to all previous tokens in $\mathbf{K}$ \& $\mathbf{V}$, resulting in per-layer KV-caches that consume considerable memory.

\textbf{Data dependent forget-gates:} To reduce large KV-cache sizes, forget-gates decay past key/value tokens, deleting sufficiently decayed entries from the KV-cache. In particular, FoX~\citep{forgetting_transformer_1, forgetting_transformer_2} expresses a forget-gate as an additive bias to the attention matrix.
%During inference, a certain structure to the decay pattern is enforced that enables sufficiently decayed key/value-tokens to be deleted from the KV-cache. 
Using terminology from~\citep{forgetting_transformer_1}, FoX computes a gate: $f_t = \sigma(\mathbf{x}_t\mathbf{w} + b) \in \mathbb{R}$ for each token in the input sequence and computes:
\begin{equation}
   \text{Softmax}(\mathbf{Q}\mathbf{K}^\top+ \mathbf{D})\mathbf{V},
\end{equation}
where $\mathbf{D}_{ij} = \sum_{l=j+1}^{i}\log(f_{l})$. These decay-values decrease monotonically from the main-diagonal of the attention matrix, 
ensuring that cache entries with near-zero gates stay near-zero in perpetuity.
% enforcing the invariant that if a query-key dot product involving key \texttt{k} is zeroed at time \texttt{t}, then all dot-products involving \texttt{k} after \texttt{t} are zeroed. 
% In doing so, 
Sufficiently decayed keys can thus be safely deleted from the KV-cache. However, 
because KV-cache entries decay in lockstep,
this decay mechanism resembles a sliding-window, an inexpressive KV-cache eviction strategy. 
% On the other hand, Universal-Attention 
UA generalizes and incorporates several such decay mechanisms, to achieve more sophisticated behavior.

\section{Universal Attention}
We propose Universal Attention (UA), a novel framework and architecture that incorporates context-aware decay into softmax attention, inspired by modern linear attention methods and the structure of the attention matrix itself. 
% This enables the model to learn sparsity patterns that allow efficient computation during both training and inference. For training, Universal Attention is \textit{hardware-aligned}, admitting a high-performance custom kernel that integrates its decay with flash attention and grouped-query attention (GQA); for inference, the decay identifies which KV-cache entries to evict, reducing KV-cache sizes.
The result is a model that learns adaptive and dynamic sparsity patterns balancing efficiency with expressivity.

\begin{figure}[!t]
    \centering
    \includegraphics[width=0.9\linewidth]{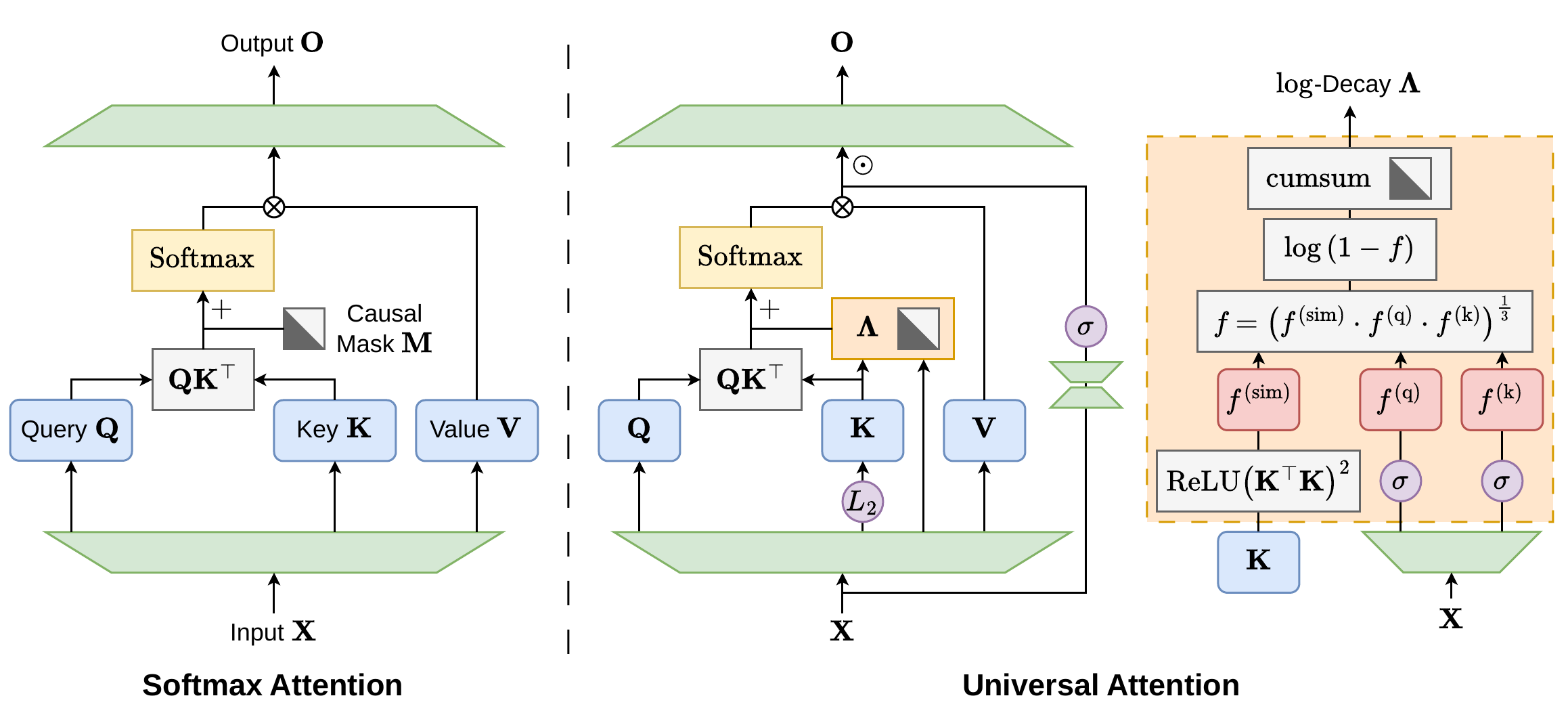}
    \caption{\textbf{Overview of Universal Attention.} (\textbf{Left}) Standard softmax attention with a fixed causal mask $ \mathbf{M}$. (\textbf{Middle}) UA augments the attention logits $\mathbf{QK}^\top$ with a learned, content-aware decay bias $\mathbf{\Lambda}$, while applying $L_2$-normalization to keys, and a low-rank sigmoid gate to the attention output. (\textbf{Right}) Construction of $\mathbf{\Lambda}$ from three decay mechanisms: similarity-based $f_\text{sim}$, query-aware $f_\text{q}$, and key-aware $f_\text{k}$. These are combined via a geometric mean, transformed to log-space, and accumulated causally to produce the decay matrix.}
    \label{fig:ua_main}
\end{figure}

Figure~\ref{fig:ua_main} depicts a UA layer, incorporating decay into softmax attention as a bias term. However, rather than relying on a single decay mechanism, UA combines multiple context-aware patterns via their geometric mean to form a unified decay rate value for each pair of positions. We choose the geometric mean over arithmetic because it preserves the veto property: if any single mechanism deems that a token should not decay, the combined decay is driven to zero.

Formally, consider the input to a UA layer as $\mathbf{X} \in \mathbb{R}^{L \times d_{model}}$, with queries, keys, and values $\mathbf{Q}, \mathbf{K}, \mathbf{V} \in \mathbb{R}^{n \times L \times d}$, where keys are L2-normalized. Universal Attention computes:
\begin{align}
    \text{Softmax}\left(\mathbf{Q}\mathbf{K}^\top + \mathbf{\Lambda}\right)\mathbf{V},\quad \mathbf{\Lambda}_{ij} := \begin{cases} \displaystyle\sum_{k=j+1}^{i} \mathbf{A}_{kj} & i \geq j \\ -\infty & i < j,\end{cases}
\end{align}
where $\mathbf{\Lambda}_{ij}$ represents the accumulated log-decay from key $j$ to query $i$: each column $j$ of $\mathbf{A}$ tracks how a single key decays across subsequent positions, and $\mathbf{\Lambda}_{ij}$ sums this decay causally, with $-\infty$ enforcing causal masking through the softmax. We operate in log-space so that multiplicative decay composes additively. The matrix $\mathbf{A}$ aggregates the decay functions in $\mathcal{F}$ via their geometric mean:
\begin{align}
    \mathbf{A} = \log\!\left(\mathbf{1} - \left(\prod_{f\in\mathcal{F}} f(\mathbf{X})\right)^{\frac{1}{|\mathcal{F}|}}\right),\label{eqn:general_ua}
\end{align}
where $\prod$ denotes the elementwise product and each context-aware decay mechanism $f\colon \mathbb{R}^{L \times d_{model}} \to [0,1]^{L \times L}$ produces a pairwise decay matrix. Below, we introduce three representative mechanisms, linked to the first- and second-order statistics of stored keys. 
% We describe these mechanisms below.

\subsection{Decay Components}

Our representative decay mechanism family $\mathcal{F}:=\left\{f^{(\text{q})}, f^{(\text{k})}, f^{(\text{sim})}\right\}$ includes (1)~a Mamba-style query-aware forget gate $f^{(\text{q})}$, (2)~a novel key-aware forget gate $f^{(\text{k})}$, and (3)~a DeltaNet-style similarity decay $f^{(\text{sim})}$. The first two mechanisms capture first-order statistics aligned to the query and key axes of the attention matrix, respectively, while the third captures second-order interaction effects between embeddings at current and past time steps.

% \davis{Subsection: decay components here. Open with a line about how we intuitively represent all available information in the 2D attention matrix: first order info linked to each axis, and second order interaction effects.}
% \ahan{addressed.}

\textbf{Query-aware forget gate $f^{(\text{q})}$:} Similar to Mamba2~\citep{mamba2} and FoX~\citep{forgetting_transformer_1}, each query position is assigned a scalar forget-gate computed as a sigmoid with a learned bias.
% , following the same parameterization as Mamba. 
% Each query's forget gate is broadcasted across all prior keys:
Since this decay value depends only on the query position, it is broadcast along the key axis to produce the corresponding pairwise decay matrix $f^{(\text{q})}$:
\begin{align}
    g^{(\text{q})}_i := \sigma\left(\mathbf{x}_i \mathbf{w}^{(\text{q})} + b^{(\text{q})}\right),\quad f^{(\text{q})}(\mathbf{X})_{ij} := g^{(\text{q})}_i,
\end{align}
%\ahan{Notationally this is a little hard for me to parse. I understand that you're doing this for the purposes of a broadcast. An alternative could be to instead straight-away do: fq = sigmoid(XW + b) - noting everything in matrix form, and using underbraces to indicate the size of the matrix. Also, notationally, for g the q is a superscript, but for f is a subscript, adding to the confusion a little bit. Let me know what you think and ok either way.}
%\haochen{Good point. I have made them all superscripts. As for using the matrix form, I tend to think element-wise is better. I'm not a huge fan of underbraces, because I don't think they are the standard approach. A mathematically sound alterantive would be writting the product of gq and qk as an outer product, which automatically gives us broadcast, but it requires a rework on all the notations.}
where $\mathbf{w}^{(\text{q})}$ and $b^{(\text{q})}$ are learned parameters. These gates accumulate causally, so that a key's importance degrades with the number of intermediate positions visited. 
This mechanism can be interpreted as each token advancing the entire cache by a variable-size time step. 

%\ahan{What is "pairwise" form here?}
%\haochen{I meant the query key pairwise, but it is indeed ambiguous. }
\textbf{Key-aware forget gate $f^{(\text{k})}$:} Symmetric to $f^{(\text{q})}$, each \textit{key} is assigned a scalar forget gate via the same sigmoid+bias parameterization, which then broadcasts over the \textit{query} dimension:
\begin{align}
    g^{(\text{k})}_j := \sigma\left(\mathbf{x}_j \mathbf{w}^{(\text{k})} + b^{(\text{k})}\right),\quad f^{(\text{k})}(\mathbf{X})_{ij} := g^{(\text{k})}_j.
\end{align}
While $f^{(\text{q})}$ represents how much history to forget, our novel $f^{(\text{k})}$ lets each key control its own intrinsic decay rate, independent of future updates. 
This allows the model to designate tokens that persist through time, or that self-prune quickly, on a case-by-case basis.
% Similarly, since $f^{(\text{k})}$ depends only on the key position, it is broadcast along the query axis to the pairwise decay matrix form. 

% \haochen{Another thing is about scalar representation vs. matrix representation. In these paragraphs it seems that I intended defined f as matrices, while in the diagram they are treated as scalar. after fixing readibility of (4) and (5), should align with that. }
% \davis{I reordered the broadcasting sentences so that they appear before the equations rather than after, which should fix readability. Not sure how to deal with matrix vs scalar - it might be fine to just abuse notation here, or we could define separate matrix $F$ vs function/scalar $f$ perhaps...}

\textbf{Similarity decay $f^{(\text{sim})}$:} 
To capture higher-order interaction effects between stored keys, we derive a pairwise decay mechanism similar to DeltaNet~\citep{delta_rule}. The DeltaNet update recurrence takes the form:
\begin{equation}
  % \mathbf{S}_t = \mathbf{S}_{t-1}\big(\mathbf{I} - \beta_t \underbrace{\mathbf{k}_t\mathbf{k}_t^\top}_{\mathbb{R}^{d \times d}}\big) + \beta_t\underbrace{\mathbf{v}_t\mathbf{k}_t^\top}_{\mathbb{R}^{d \times d}}
  \mathbf{S}_t = \mathbf{S}_{t-1}\big(\mathbf{I} - \beta_t \mathbf{k}_t\mathbf{k}_t^\top\big) + \beta_t\mathbf{v}_t\mathbf{k}_t^\top
  % , \quad \mathbf{o}_t = \mathbf{S}_t\mathbf{q}_t
\end{equation}

The state $\mathbf{S}_{t}$ accumulates key/value outer products with step size $\beta_t$, and so we can unroll the prior state $\mathbf{S}_{t-1}:=\sum_{i=0}^{t-1}\mathbf{v}_i'\mathbf{k}_i'^{\top}$, where $\textbf{v}_i',\textbf{k}_i'$ represent $\textbf{v}_i,\textbf{k}_i$ with changes absorbed from subsequent state updates. DeltaNet sets $\textbf{k}_i':=\textbf{k}_i$ and absorbs all changes into $\textbf{v}_i'$, but an equivalent formulation sets $\textbf{v}_i':=\textbf{v}_i$ and absorbs all changes into $\textbf{k}_i'$:

\begin{equation}
    \mathbf{S}_{t-1}\big(\mathbf{I} - \beta_t \mathbf{k}_t\mathbf{k}_t^\top\big) 
    = \sum_{i=0}^{t-1}\mathbf{v}_i\textbf{k}_i'^\top \big(\mathbf{I} - \beta_t \mathbf{k}_t\mathbf{k}_t^\top\big) 
    = \sum_{i=0}^{t-1}\mathbf{v}_i \big( \textbf{k}_i' - \beta_t\textbf{k}_i'\textbf{k}_t^\top\textbf{k}_t\big)^\top
\end{equation}

For $L_2$-normalized $\textbf{k}$, the right-hand term inside the summation represents the projection of $\textbf{k}_i'$ onto $\textbf{k}_t$, scaled by $\beta_t\in[0,1]$. Thus we can interpret the DeltaNet update as a \textit{partial orthogonalization} of all prior keys against the newest key entering the cache. Over successive rounds of orthogonalization, the magnitude of $\textbf{k}_i'$ decays to zero, slowly when it is not aligned to each new $\textbf{k}_t$, and quickly when it is.

Unfortunately, this decay mechanism is not UA-compatible as-is, because dynamically shifting $\textbf{k}$ both 1) interferes with RoPE, and 2) requires overwriting the entire key cache at each time step, which is computationally prohibitive.
We propose a simple modification that transforms DeltaNet decay into an attention-bias-friendly form:
\begin{align}
    \mathbf{k}_i' - \beta_t\mathbf{k}_i'^\top\mathbf{k}_{t}\mathbf{k}_{t}
    \quad\Rightarrow\quad
    \mathbf{k}_i' - \beta_t\mathbf{k}_i'^\top\mathbf{k}_{t}{\color{blue}\mathbf{k}_i'}\ =\ (1-\beta_t\mathbf{k}_i'^\top\mathbf{k}_{t})\textbf{k}_i'
\end{align}
% We propose a simple modification on the orthogonalize step that preserves direction while reducing magnitude based on key similarity:
% \begin{align}
%     \mathbf{k}_{t-1}^\top - \mathbf{k}_{t-1}^\top\mathbf{k}_{t}\mathbf{k}_{t}^\top\quad\Rightarrow\quad\mathbf{k}_{t-1}^\top - \mathbf{k}_{t-1}^\top\mathbf{k}_{t}\mathbf{k}^\top_{\color{blue}t-1}.
% \end{align}
% \haochen{I feel like a highlight using a different color is needed here, otherwise reviewers that are familiar with GDN will question this. }
By replacing $\mathbf{k}_t$ with $\mathbf{k}_i'$ in the rightmost term, the update reduces to 
% scaling $\mathbf{k}_{t-1}^\top$ by $(1 - \mathbf{k}_{t-1}^\top\mathbf{k}_t)$, 
a multiplicative decay factor determined by key similarity. The decay factor is thus $\beta_t\mathbf{k}_i'^\top\mathbf{k}_t$, which serves as the basis for $f^{(\text{sim})}$. We let $f^{(\text{query})}$ absorb $\beta_t$ and define $f^{(\text{sim})}$ based solely on the dot product:
% We apply ReLU so that dissimilar keys never amplify each other, ensuring non-negative decay per key, and square the similarity to concentrate decay on near-duplicate keys, yielding the pairwise decay: 
\begin{align}
    f^{(\text{sim})}(\mathbf{X})_{ij} := \mathrm{ReLU}(\mathbf{k}_i^\top \mathbf{k}_j)^2
\end{align}
Since keys are $L_2$-normalized, $\mathbf{k}_i^\top\mathbf{k}_j \in [-1, 1]$, and squared ReLU ensures a gate value in $[0,1]$ as desired. The result is a conservative decay mechanism biased heavily toward zero, but still capable of eliminating a key/value pair when a new entry is perfectly aligned.

While we focus on these three representative mechanisms, the framework is extensible: additional decay functions can be incorporated into $\mathcal{F}$ as new advances emerge in linear attention --- hence the \textit{universal} nature of our method.

\subsection{Layer Design and Algorithm}
% \paragraph{Instantiated form.} 
Plugging $\mathcal{F} = \{f^{(\text{q})}, f^{(\text{k})}, f^{(\text{sim})}\}$ into the general framework~\ref{eqn:general_ua}, the decay matrix becomes:
\begin{align}
\label{eq:ua_final}
    \mathbf{A}_{ij} = \log\!\left(1 - \Big(g^{(\text{q})}_i \cdot g^{(\text{k})}_j \cdot \mathrm{ReLU}(\mathbf{k}_i^\top \mathbf{k}_j)^2\Big)^{\frac{1}{3}}\right).
\end{align}
The accumulated log-decay $\mathbf{\Lambda}_{ij} = \sum_{k=j+1}^{i} \mathbf{A}_{kj}$ can then be added to the pre-softmax logits of our attention layer, depicted in Fig.~\ref{fig:ua_main}. 
% In addition to the decay mechanisms, 
We also take architectural inspiration from linear attention models, which often apply additional channel-wise sigmoid gates to the intermediate representation~\citep{mamba2,gated_delta_net,hyena}. We adopt analogous gates for attention layers~\citep{gated_attention}, forming unique per-channel gates via low-rank projection~\citep{kimi_linear}. 
With rank equal to head width, these gates add a single attention head's worth of extra parameters per layer, a negligible quantity with outsize impact on performance. By reducing reliance on sink tokens~\citep{gated_attention}, channel gating allows the decay mechanisms greater freedom to learn expressive patterns. We ablate this component and compare to both gated and vanilla attention baselines in Section~\ref{sec:main_results}.

\textbf{Inference pruning:} Beyond shaping attention, UA’s learned decay also provides a natural signal for inference-time KV-cache pruning. 
% The three decay components capture complementary aspects of token importance, which together define an importance score for each cached token. 
The log-decay matrix $\Lambda$ is implemented as a bias on attention logits, but it can also be interpreted, through elementwise exponentiation, as a set of token-wise masks applied to the logits post-softmax:
\begin{equation}
    \text{Softmax}\left(\mathbf{Q}\mathbf{K}^\top + \mathbf{\Lambda}\right)_{ij}
    = \frac{\text{e}^{(\mathbf{Q}_i\mathbf{K}_j^\top+\mathbf{\Lambda}_{ij})}}{\sum_{k=0}^j\text{e}^{(\mathbf{Q}_i\mathbf{K}_k^\top+\mathbf{\Lambda}_{ik})}}
    = \frac{\text{e}^{\mathbf{Q}_i\mathbf{K}_j^\top}\exp(\mathbf{\Lambda})_{ij}}{\sum_{k=0}^j \text{e}^{\mathbf{Q}_i\mathbf{K}_k^\top}\exp(\mathbf{\Lambda})_{ik}}
\end{equation}
Mask values in $\exp(\mathbf{\Lambda})$ start at 1 and decay to 0 asymptotically, providing a natural pruning criterion: 
% . Just as $\Delta$ accumulates log-gate biases additively, $exp(\Lambda)$ represents the cumulative product of those same forget gates, starting at 1 and decaying to 0.
% This yields a simple online pruning rule: 
during inference, maintain a scalar bias $\lambda_i$ for each key/value pair representing the corresponding entry of $\mathbf{\Lambda}$, continuously update $\lambda_i$ using Eq.~\ref{eq:ua_final}, and evict any key-value pair whose effective mask $e^{\lambda_i}$ falls below a chosen threshold $p\in[0,1]$. 
% Since decay is always non-negative, the corresponding importance score decreases monotonically, 
Decay bias updates in $\textbf{A}$ are negative by construction, so $\lambda_i$ always decreases, 
thus a token with $\lambda_i<\log p$ will not return to importance later. Although simple, this rule fully exploits the learned decay and lets the model itself decide, on a head-by-head basis, which and how many tokens should be retained. More sophisticated pruning algorithms are left to future work.
%% Older paragraphs. %%
%During inference, Universal-Attention prunes the KV-cache using the decay weights during each decode step. Specifically, once the decay weights of a token go below $\delta$ (for some threshold value $\delta$), we delete it from the KV-cache. Since decay-weights only decrease as inference proceeds, we are guaranteed that evicted tokens will not impact the attention calculation for a properly chosen threshold. 

\textbf{Implementation:} We codesign UA to be \textit{hardware-aligned} -- interoperating with GQA~\citep{grouped_query_attention} and FlashAttention~\citep{fa,fa2,fa3,fa4}, as current kernels do not support both GQA and float-valued biases. We implement UA as two kernels, separately computing (1) the decay-mask, and (2) the attention-output. Additionally, we apply Activation-Checkpointing~\citep{activation-checkpointing} to mitigate storing the \texttt{O(L$^2$)} sized decay-mask \textit{per-layer} in device memory - which would otherwise be required to compute gradients in the backwards pass. Our custom implementation boosts end-to-end training throughput by 1.13$\times$ compared to PyTorch-2. We explain our kernel implementations and end-to-end pipeline in detail in Appendix~\ref{app:kernel-design}.

\section{Evaluation}

\textbf{Setup:} We evaluate Universal Attention in a 1B-parameter transformer following the Llama3 architecture~\citep{llama3}, which uses RoPE~\citep{rope} and Grouped Query Attention~\citep{grouped_query_attention}.
Models are trained for 50B tokens on Dolma v1.7 data~\citep{dolma}, reweighted per~\cite{bamba}. 
Model and training details and hyperparameters are provided in Appendix~\ref{app:experiment_setup}.
% , and gated attention~\citep{gated_attention}. 
For long-context inputs, we apply YaRN~\citep{yarn} as a tuning-free adjustment to RoPE. We report results at two pruning thresholds $p=(.05,.15)$, to illustrate performance tradeoffs at conservative/aggressive levels of compression. For UA models, we additionally prune in chunks of 128 tokens during prefill, rather than running the whole input and only pruning afterwards.

% \davis{Should we mention that we prune the prefill in chunks here?}

\textbf{Baselines:} We compare against both linear- and softmax-attention baselines: Mamba2~\citep{mamba2}, Mamba2-hybrid~\citep{nemotron-h}, Gated DeltaNet (GDN) and its hybrid~\citep{gated_delta_net}, as well as standard Llama3 and gated-attention Llama3-G. All baselines are trained identically to UA. 

For context extension, we apply YaRN~\citep{yarn} only to the Llama3-based models, where RoPE rescaling is a standard extension recipe. For Mamba2-hybrid, we follow Nemotron-H and keep its native RoPE-free design in the attention layers. {\color{edits}For GDN and its hybrid, we retain the base RoPE setup in the main comparison to avoid introducing an additional positional intervention. Appendix~\ref{app:ruler_16k} further evaluates GDN-H with YaRN at 8K and 16K, where it improves long-context performance but does not change the overall comparison with UA.}

To compare KV cache compression efficiency, we also apply H$_2$O~\citep{h2o}, SnapKV~\citep{snapkv}, and their pyramid variants~\citep{pyramidkv} to the vanilla Llama3 model. {\color{edits}We also include KeyDiff~\citep{keydiff} as an additional baseline using the publicly available kvpress~\citep{kvpress} implementation.} To match the compression rate of UA, we aggressively set these methods to keep only 10\% of the KV-cache. This strong suite of baselines thus encompasses a range of complementary recent approaches to reducing memory overhead, as well as two full-size oracles. Further details are provided in Appendix~\ref{app:experiment_setup}.

\textbf{Benchmarks:} We evaluate the models' performance on two sets of benchmarks: information retrieval, including LongBench-E~\citep{longbench} and RULER~\citep{ruler}, and common-sense reasoning tasks, following \cite{gated_delta_net}. For each task, we track the size of the stored states and KV-caches (``effective cache size'') using their actual sizes in bytes.

\subsection{Downstream Performance}
\label{sec:main_results}

\textbf{Information retrieval:} Table~\ref{tab:ruler_longbench} reports results on inputs within the training sequence length, together with effective cache size, for RULER (4K) and LongBench-E (0-4K). On RULER, 
UA delivers significant improvement, even over unpruned Llama3 oracle models, claiming the highest and second-highest average scores at its two pruning thresholds. 
This improvement is mostly fueled by gains in tracking and retrieval tasks, though performance on aggregation suffers.
% Universal Attention delivers the strongest overall performance: $p{=}0.05$ achieves the best average score (50.3) as well as the best retrieval (68.4) and tracking (36.4), while $p{=}0.15$ attains a nearly identical average (49.3) with only 12.9 MB of cache. 
% Both UA variants outperform the strongest quadratic and sub-quadratic baselines at substantially lower memory cost. 
Aggregation is a common bottleneck for KV-cache eviction methods, since it requires combining information from broadly distributed context. UA shares this weakness, but unlike SnapKV,  H$_2$O, {\color{edits} and KeyDiff, its substantial degradation is concentrated in aggregation}. 
On LongBench-E, the best average is obtained by Mamba2-H, yet UA remains highly competitive under a much tighter cache budget: $p{=}0.15$ 
uses under half the memory of Mamba2-H, and exceeds both Llama3 and Llama3-G scores with only a fraction of their cache size.
This favorable comparison to full-cache Llama3 also persists at 3B model scale, with results in Appendix~\ref{app:3B_model_results}.
% reaches 18.0 with 14.4 MB, exceeding both Llama3 and Llama3-Gated while using only a small fraction of their cache. 
Overall, UA offers superior memory-performance tradeoffs.
% Overall, the table shows that UA offers one of the best performance--memory trade-offs controlled by the pruning threshold $p$.
% \input{tables/table_longbench}
\begin{table}[!ht]
\centering
\caption{RULER (4k sequence length) and LongBench-E (0-4k sequence length split). 
% KV cache size reports the average token count retained across all layers.
}
\label{tab:ruler_longbench}
\resizebox{0.9\textwidth}{!}{
\scriptsize
\begin{tabular}{lcccccc|cc}
\toprule
&\multicolumn{6}{c|}{\textbf{RULER (4k)}}& \multicolumn{2}{c}{\textbf{LongBench-E (0-4k)}}\\
\cmidrule(lr){2-7}\cmidrule(lr){8-9}
\textbf{Model} & \textbf{Retrieval} & \textbf{Tracking} & \textbf{Aggregation} & \textbf{QA} & \textbf{Avg.} & \textbf{Cache (MB)}  &\textbf{Avg.} & \textbf{Cache (MB)} \\
\midrule\midrule
\multicolumn{9}{l}{\textit{Sub-quadratic models}} \\
\quad Mamba2 & 31.0 & 0.8 & 18.1 & 20.0 & 25.0 & 20 &12.7	& 20.0\\
\quad Mamba2-H & 57.2 & 3.2 & \underline{20.9} & \underline{29.7} & 43.2 & 35.5 &\textbf{18.9}	&31.7\\
\quad GDN & 37.7 & 2.6 & 18.1 & 18.4 & 29.0 & 32 &14.9	&32.0\\
\quad GDN-H & 59.9 & 4.4 & 19.6 & 24.9 & 44.1 & 46.0 &15.7	&42.2\\
\midrule
\multicolumn{7}{l}{\textit{Quadratic models}} \\
\quad Llama3 & 54.2 & 8.4 & 13.6 & \textbf{30.4} & 40.8 & 143.7 &16.4	&113.3 \\
\quad \quad + SnapKV & 21.7 & 13.4 & 7.5 & 25.6 & 19.5 &  14.4 & 12.9	& 11.3\\
\quad \quad + Pyramid SnapKV & 21.2 & 20.3 & 6.2 & 25.1 & 19.4 & 14.4 & 12.9	& 11.3\\
\quad \quad + H$_2$O & 4.2 & 0.2 & 10.7 & 19.9 & 7.3 & 14.4 & 13.0	& 11.3\\
\quad \quad + Pyramid H$_2$O & 4.3 & 0.2 & 9.7 & 20.0 & 7.2 & 14.4 & 12.6	& 11.3\\
\quad \quad + KeyDiff ($K=10\%L$) & 33.8	& 24.7	& 10.2	& 24.3	& 28.0	& 14.4	& 15.2	& 11.3\\
\quad Llama3-G & 53.7 & 16.3 & \textbf{21.4} & 28.3 & 41.9 & 143.7 & 17.9	& 113.3\\
\midrule
\multicolumn{9}{l}{\cellcolor{customblue}\textit{Universal Attention}} \\
\quad \cellcolor{customblue}\textbf{p=0.05} & \cellcolor{customblue}\textbf{68.4}& \cellcolor{customblue}\textbf{36.4}& \cellcolor{customblue}5.7& \cellcolor{customblue}29.5& \cellcolor{customblue}\textbf{50.3}& \cellcolor{customblue}21.4 &\cellcolor{customblue}17.9	&\cellcolor{customblue}20.2\\
\quad \cellcolor{customblue}\textbf{p=0.15} & \cellcolor{customblue}\underline{67.3} & \cellcolor{customblue}\underline{30.6} & \cellcolor{customblue}6.9 & \cellcolor{customblue}29.1 & \cellcolor{customblue}\underline{49.3} & \cellcolor{customblue}12.9 &
\cellcolor{customblue}\underline{18.0} & 
\cellcolor{customblue}14.4\\
\midrule\bottomrule
\end{tabular}
}
\end{table}

\textbf{Common-sense reasoning:} Table~\ref{tab:short_context} presents language modeling and zero-shot reasoning results at a context length of 4k. 
Due to benchmark setup, H$_2$O, SnapKV, and KeyDiff only prune their caches \textit{after} producing output logits from full-cache attention, and so are omitted from comparison.
% Although H$_2$O and SnapKV prune the KV cache during prefilling, their output logits are still computed from full-cache attention, with eviction applied only afterward based on the resulting attention scores. They therefore do not change the outputs for these tasks and are omitted from comparison. 
% Universal Attention achieves the lowest perplexity on LAMBADA (10.61), outperforming all baselines including recurrent models and hybrids. On zero-shot reasoning, Universal Attention attains the highest average accuracy (50.88), with notable gains on LAMBADA accuracy (+3.2 over the next best), PIQA, and ARC-easy. 
UA achieves competitive-to-superior perplexity and reasoning scores, with notable gains in LAMBADA accuracy, PIQA, and ARC-easy, despite aggressive pruning to only 38.7 KV-cache tokens on average.
{\color{edits}
While the absolute zero-shot reasoning gains are modest, these results provide strong evidence that UA preserves model quality at substantially lower cache cost: UA matches or slightly exceeds Llama3 in average reasoning accuracy (50.74/50.88 vs. 50.28) while reducing cache from 2.6 MB to 1.9/1.5 MB, and uses roughly 10--13$\times$ less cache than Mamba2 and 17--21$\times$ less than GDN while maintaining comparable or better accuracy. 
}
Note that due to short input sequence length, the Mamba2 and GDN linear attention architectures consume more memory than the corresponding full attention models, and UA compression levels are less dramatic in comparison, though easily still the lowest effective cache size overall. {\color{edits} This demonstrates that the learned decay preserves short-context quality and improves the memory--quality tradeoff, while potentially acting as an implicit regularizer for perplexity and downstream task performance.}

\begin{table}[!ht]
\centering
\caption{Language modeling perplexity and zero-shot common-sense reasoning accuracy.}
\label{tab:short_context}
\resizebox{0.95\textwidth}{!}{
\scriptsize
\begin{tabular}{lcc|ccccccccccc}
\toprule
 & \multicolumn{2}{c|}{\textbf{Perplexity ($\downarrow$)}} & \multicolumn{10}{c}{\textbf{Zero-Shot Common-Sense Reasoning (acc $\uparrow$)}} \\
\cmidrule(lr){2-3} \cmidrule(lr){4-13}
\textbf{Model} & \textbf{Wiki.} & \textbf{LMB.} & \textbf{LMB.} & \textbf{PIQA} & \textbf{Hella.} & \textbf{Wino.} & \textbf{ARC-e} & \textbf{ARC-c} & \textbf{SIQA} & \textbf{BoolQ} & \textbf{Avg.} & \textbf{Cache (MB)} \\
\midrule\midrule
Mamba2            & 19.36 & 12.20 & 47.20 & \underline{72.20} & \underline{50.94} & 54.93 & 55.47 & 27.39 & \textbf{41.25} & 54.86 & 50.53 & 20 \\
Mamba2-H     & \textbf{18.54} & 10.97 & 49.62 & 71.71 & \textbf{51.49} & \textbf{55.41} & 56.36 & 26.54 & \underline{40.02} & 55.11 & 50.78 & 17.8 \\
GDN    & 19.29 & 11.42 & 48.65 & 71.49 & 50.25 & 54.70 & \textbf{57.32} & 27.30 & 39.46 & 57.65 & \underline{50.85} & 32 \\
GDN-H        & 18.90 & 11.21 & 49.27 & 71.87 & 50.61 & 53.35 & 55.81 & \underline{27.82} & 39.51 & 56.94 & 50.65 & 28.3 \\
Llama3            & 19.40 & 11.95 & 47.89 & 71.71 & 49.65 & 53.99 & 54.42 & 25.68 & 39.61 & \textbf{59.30} & 50.28 & 2.6 \\
Llama3-G      & 19.38 & 12.28 & 47.23 & 71.87 & 49.58 & \underline{55.17} & 54.21 & \textbf{27.82} & 39.82 & \underline{58.84} & 50.57 & 2.6 \\
\cellcolor{customblue}\textbf{UA, p=0.05} & \cellcolor{customblue}\underline{18.67} & \cellcolor{customblue}\textbf{10.52} & \cellcolor{customblue}\underline{50.92} & \cellcolor{customblue}72.14 & \cellcolor{customblue}50.89 & \cellcolor{customblue}53.51 & \cellcolor{customblue}\underline{56.82} & \cellcolor{customblue}25.68 & \cellcolor{customblue}39.76 & \cellcolor{customblue}56.18 & \cellcolor{customblue}50.74 & \cellcolor{customblue} 1.9 \\
\cellcolor{customblue}\textbf{UA, p=0.15} & \cellcolor{customblue}19.03 & \cellcolor{customblue}\underline{10.61} & \cellcolor{customblue}\textbf{51.12} & \cellcolor{customblue}\textbf{72.36} & \cellcolor{customblue}50.93 & \cellcolor{customblue}53.12 & \cellcolor{customblue}56.69 & \cellcolor{customblue}25.94 & \cellcolor{customblue}39.87 & \cellcolor{customblue}57.03 & \cellcolor{customblue}\textbf{50.88} & \cellcolor{customblue} 1.5 \\
\midrule\bottomrule
\end{tabular}
}
\end{table}

\subsection{Context Length Extension}\label{sec:cle}
\begin{figure}
    \centering
    \includegraphics[width=0.95\linewidth]{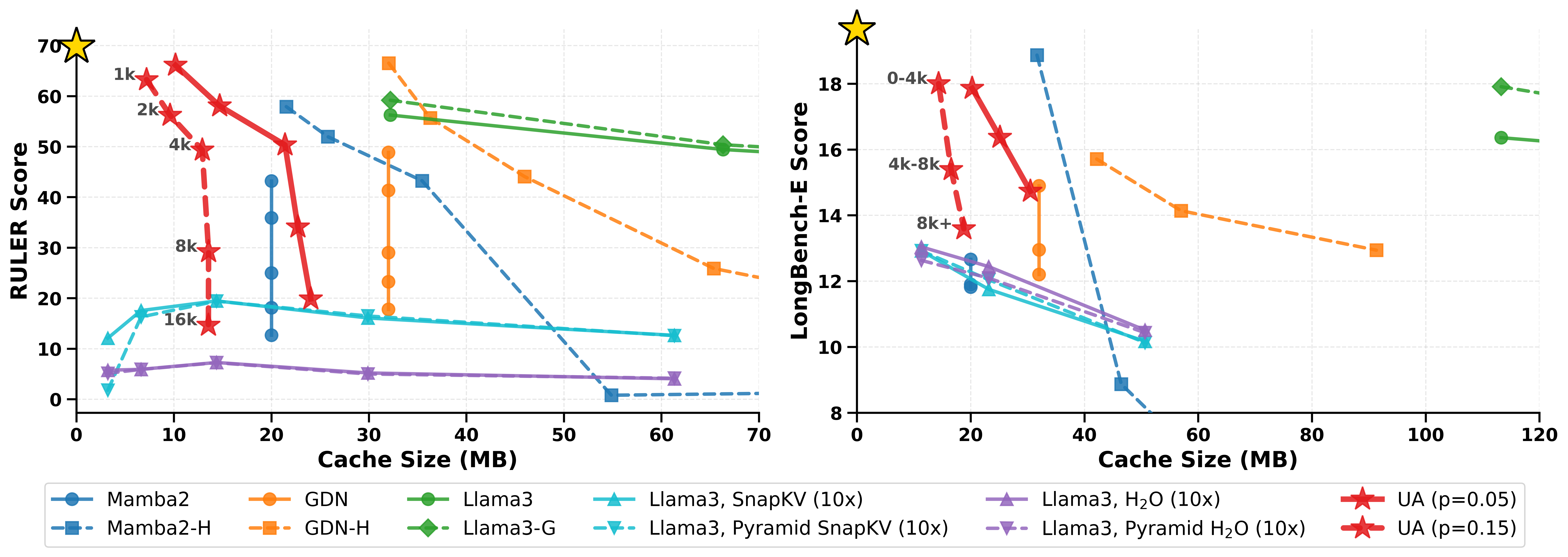}
    \caption{Score versus effective cache size trade-off on RULER (\textbf{left}) and LongBench-E (\textbf{right}). {\color{edits}Each line traces a model across increasing sequence lengths from 1K to 16K; thus, moving along a curve changes both cache size and task difficulty.} The gold star ($\bigstar$) denotes the ideal operating point of high score and low overhead. 
    % configurations closer to it achieve a better memory–performance trade-off.
    }
    \label{fig:RULER_LongBench}
\end{figure}

To better understand how the memory-performance trade-off evolves with context length, Figure~\ref{fig:RULER_LongBench} plots average score against effective cache size for RULER (left) and LongBench-E (right). {\color{edits} Each curve traces a model across increasing sequence lengths from 1K to 16K, so both cache size and task difficulty vary along the curve. Consequently, a lower score at a larger cache size does not imply that additional cache is harmful, but simply reflects evaluation at a longer and more challenging context. Better configurations at a given context length lie toward the upper-left,} and UA consistently traces the strongest frontier as sequence length increases.

The two UA variants expose a clear trade-off within the method itself: $p{=}0.05$ preserves higher scores, while $p{=}0.15$ achieves more aggressive compression. Unlike full-attention baselines such as Llama3 and Llama3-G, whose cache grows linearly with context length and quickly exceeds the plotting range, UA keeps cache usage under 25 MB even at the longest contexts shown. Relative to Llama3, UA with $p{=}0.15$ achieves $4.5\times$--$45.3\times$ cache reduction on RULER and $7.9\times$--$26.9\times$ on LongBench-E. The gap widens at longer contexts because UA's cache plateaus, while Llama3 continues to grow. By contrast, SnapKV and H$_2$O reduce cache at a larger accuracy cost, while hybrid baselines grow the cache with diminishing returns. The full breakdown of benchmark scores by task and sequence length is provided in Appendix~\ref{app:ruler_16k}.

\subsection{Empirical Analysis}
\label{sec:empirical}
\begin{figure}[!t]
    \centering
    \includegraphics[width=0.9\linewidth]{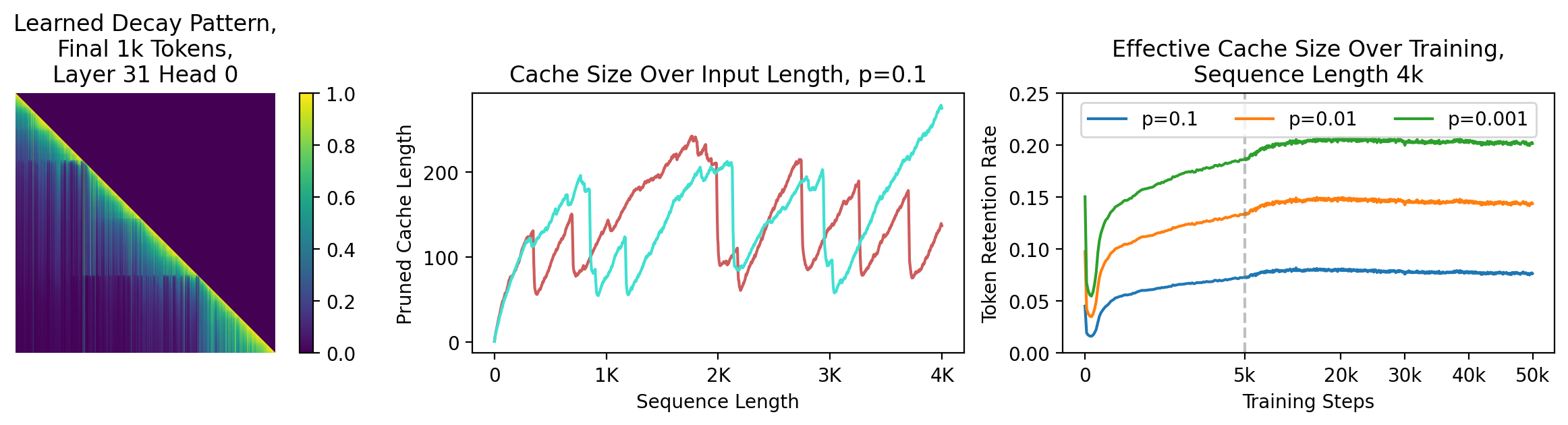}
    \caption{
    (\textbf{Left}) One example head's learned pattern of accumulated forget gates $exp(\mathbf{\Lambda})$, exhibiting behaviors from multiple decay mechanisms.
    (\textbf{Middle}) Progressive UA cache size over two different input sequences. Pruning is highly dynamic and input-adaptive, with spontaneously learned growth-and-reset behaviors.
    (\textbf{Right}) Evolution of effective cache size over the course of training, with an extreme initial dip and gradual recovery.
    }
    \label{fig:ua_empirical}
\end{figure}

We examine UA's actual learned patterns, uncovering several interesting and spontaneous behaviors. Fig.~\ref{fig:ua_empirical} (left) shows an example learned decay pattern, expressed as the set of accumulated masks 
$\exp{(\mathbf{\Lambda})}$.
% applied to the softmax scores -- or equivalently, as the elementwise-exponent of the UA bias term $\mathbf{\Lambda}$. 
Attention logits are not pictured. We can isolate contributions from each of our three decay mechanisms: vertical stripes represent tokens with low key-linked decay $f^{(\text{k})}$, horizontal lines represent selective filtering through similarity-based decay $f^{(\text{sim})}$, and progressive query-linked decay $f^{(\text{q})}$ causes entries to fade with distance from the diagonal. UA's mixture of mechanisms appears to be expressive.

Dynamic pruning is illustrated in Fig.~\ref{fig:ua_empirical} (middle), plotting average cache size across heads as a function of input length. While cache length is small relative to input length, the exact size varies, both over time and between different inputs. The model learns to be naturally parsimonious, accumulating information when it is needed, and pruning aggressively once it is not. This ability to discard unnecessary information could explain the consistent improvement of UA over the oracle Llama3 baselines in Section~\ref{sec:main_results}.

This notion of learned frugality is reinforced in Fig~\ref{fig:ua_empirical} (right), which plots the effective KV-cache size during training. 
Here we measure effective size as the percentage of cache entries above $p$ at the final token position / in the final row of the attention matrix.
Caches immediately sparsify at the beginning, then the model gradually re-incorporates past information as its language modeling capabilities improve. Eventually, cache sizes stabilize and even fall gently over the last half of training, with the model naturally becoming more efficient. This suggests that longer training regimes could produce further compression.

\subsection{Ablations}\label{sec:ablations}

\textbf{UA decay mechanisms:}
We ablate UA's decay mechanisms using models trained on 10B tokens instead of 50B, and report averaged RULER \texttt{S1}, \texttt{S2}, \& \texttt{S3} scores across sequence lengths. Table~\ref{tab:ablation} (left) shows that all components contribute to performance, supporting our ``universal'' approach of combining multiple decay mechanisms. Interestingly, removing the most well-studied, Mamba-style query-aware decay $f^{(\text{q})}$ causes the smallest degradation, while the other decay mechanisms provide substantial complementary benefits. Appendix~\ref{app:add_ablation} provides additional architectural ablations, including each decay mechanism in isolation, Qwen3-style normalized queries and keys, and comparisons with FoX and sliding-window attention.

{\color{edits}
\textbf{Pruning threshold:}
We ablate different pruning thresholds $p$ on our 50B-token model in Table~\ref{tab:ablation} (right), evaluating LongBench-E with inputs truncated to 16K and perplexity on held-out natural-language validation data matching the training distribution. We report both the average compression ratio across KV heads and the maximum compression achieved on a single sequence. UA degrades gracefully as compression increases: at $p{=}0.15$, it achieves $21.4\times$ average compression on LongBench-E and $14.9\times$ on held-out validation data, while changing the LongBench-E score and validation perplexity by less than one point each. Because the learned decay patterns vary across models and the compression--performance trade-off changes smoothly rather than exhibiting a natural cutoff, there is no single universally optimal threshold. We therefore use $p{=}0.05$ and $p{=}0.15$ as representative operating points for gentler and more aggressive pruning, respectively, while allowing the threshold to be selected according to the desired memory--quality trade-off.
}

\begin{table}[!ht]
\centering
\scriptsize
\caption{\textbf{(Left)}: Impact of UA components on downstream retrieval tasks. \textbf{(Right)}: Impact of pruning threshold, with performance/compression tradeoff.} 
\label{tab:ablation}
\setlength{\tabcolsep}{3.5pt}
\begin{subtable}[t]{0.38\columnwidth}
\centering
%\caption{Impact of UA components on downstream retrieval performance}
% \label{tab:ablation}
\vspace{2pt}
\begin{tabular}{lccc}
\toprule
 & \multicolumn{3}{c}{\textbf{RULER S1, S2, S3 (\% $\uparrow$)}} \\
\cmidrule(lr){2-4}
\textbf{Method} & \textbf{4k} & \textbf{8k} & \textbf{16k}\\
\midrule
\textbf{Universal Attention} & \cellcolor{customblue}\textbf{99.60} & \cellcolor{customblue}\textbf{94.20} & \cellcolor{customblue}\textbf{58.40} \\
\quad -- Query-aware $f^{(\text{q})}$ & --0.27 & --5.67 & --3.47 \\
\quad -- Key-aware $f^{(\text{k})}$ & --23.20 & --11.00 & --36.20 \\
\quad -- Similarity $f^{(\text{sim})}$ & --2.33 & --34.27 & --31.76 \\
\quad -- RoPE \& YaRN & --9.07 & --11.20 & --9.80 \\
\quad -- Gated attention & --2.93 & --20.07 & --8.13 \\
\bottomrule
\end{tabular}
\end{subtable}%
\hspace{10pt}%
\begin{subtable}[t]{0.58\columnwidth}
\centering
%\caption{Impact of pruning threshold, with performance/compression tradeoff}
% \label{tab:LongBench-sparsity}
\vspace{2pt}
\begin{tabular}{lcccccc}
\toprule
 & \multicolumn{3}{c}{\textbf{LongBench-E (16k)}} & \multicolumn{3}{c}{\textbf{Held-Out Val.\ (4k)}} \\
\cmidrule(lr){2-4} \cmidrule(lr){5-7}
$p$ & \textbf{Avg.} & \textbf{Avg.\ Comp.} & \textbf{Max Comp.} & \textbf{PPL} & \textbf{Avg.\ Comp.} & \textbf{Max Comp.} \\
\midrule
0.00 & 14.05 & 1.00 & 1.00 & 10.48 & 1.00 & 1.00 \\
\cellcolor{customblue}\textbf{0.05} & \cellcolor{customblue}\textbf{13.85} & \cellcolor{customblue}\textbf{12.94} & \cellcolor{customblue}\textbf{56.16} & \cellcolor{customblue}\textbf{10.54} & \cellcolor{customblue}\textbf{9.99} & \cellcolor{customblue}\textbf{50.88} \\
0.10 & 13.71 & 17.01 & 107.9 & 10.61 & 12.50 & 73.14 \\
\cellcolor{customblue}\textbf{0.15} & \cellcolor{customblue}\textbf{13.66} & \cellcolor{customblue}\textbf{21.38} & \cellcolor{customblue}\textbf{113.95} & \cellcolor{customblue}\textbf{10.71} & \cellcolor{customblue}\textbf{14.94} & \cellcolor{customblue}\textbf{95.26} \\
0.20 & 13.04 & 26.32 & 144.24 & 10.86 & 17.20 & 122.27 \\
0.25 & 12.24 & 31.95 & 174.07 & 11.09 & 19.61 & 151.70 \\
\bottomrule
\end{tabular}
\end{subtable}
\end{table}

%Ablate each mechanism, and rope. Maybe gating too? Mamba-style appears to be the least important!
%
%Ablate pruning threshold.

{\color{edits}
\section{Limitations}
\textbf{Aggregation tasks:}
Universal Attention is comparatively weaker on aggregation tasks, which often require combining weak evidence distributed across many positions rather than preserving a sparse set of individually salient tokens. This contrasts with UA's strengths on retrieval, tracking, and QA, suggesting a trade-off in the types of long-range information emphasized by different architectures. Our hybrid results in Appendix~\ref{app:hybrid_results} further show that this weakness can be mitigated by combining UA with complementary memory mechanisms, making UA suitable as a cache-efficient attention component within broader long-context architectures when stronger aggregation capability is required.

\textbf{Additional computation:}
UA introduces additional computation for constructing its learned decay and pruning masks. Although this enables adaptive, content-dependent retention, reducing the associated overhead remains an important systems optimization target.

\textbf{Training from scratch:} We evaluate UA as a novel architecture, but given that it generalizes standard softmax attention, a natural question is whether existing models can be adapted to UA, rather than having to pretrain it completely from scratch. Early results in Appendix~\ref{app:post_train_results} suggest that this is possible, but a significant gap remains compared to fully-pretrained UA. Closing this gap would greatly increase the practical feasibility of UA's deployment.

\textbf{Inference optimization:}
While preliminary inference experiments in Appendix~\ref{app:latency} demonstrate that UA's reduced cache can translate into lower decode latency, the current implementation does not include a fully fused, inference-optimized dynamic cache manager. Efficiently exploiting dynamically sparse KV caches requires specialized sparse kernels and systems support~\citep{flashinfer}, and developing such an implementation remains an important direction for future work.
}

\section{Conclusion}
%Discuss memory savings not yet translating to latency/throughput improvements, due to nature of inference bottlenecks (prefill vs decode), unoptimized inference implementation, and head dynamics. Perhaps discuss chunked pruned prefill as future work (even if we are pruning that way currently).
%
%But focus on memory savings, dynamic cache size and downstream performance overall. Consider splitting off above into a separate ``Discussion and Future Work'' section. 

We present Universal Attention, a technique that generalizes existing decay mechanisms, introduces a novel decay mechanism, and combines these with RoPE and softmax attention into a single, unified layer. 
% that serves as a natural adaptive pruning criterion. 
UA's expressive decay mechanism provides a natural adaptive pruning criterion, yielding superior memory-performance tradeoffs, improving downstream performance with lower memory overhead than even $O(1)$ linear attention models. Our results demonstrate that trainable self-pruning architectures provide a practical avenue to mitigate large memory overhead during inference.

%\ahan{creating comment for a standard terminology set: 
%pruning threshold - $p$. Putting $p$ here since most of the text uses it over $\delta$. \\
%Universal Attention - UA \\
%FlashAttention \\
%Grouped Query Attention \\
%(learned) decay - $\Lambda$ \\ 
%RoPE \\
%Gated Delta Networks - GDN \\
%RULER \\
%LongBench-E \\
%Softmax \\
%KV-cache(s) \\
%hardware-aware. \\
%query-aware decay - $f^{(q)}$ \\
%key-aware decay - $f^{(k)}$ \\
%similarity decay - $f^{(sim)}$ \\
%FoX \\
%PaTH \\
%H$_2$O - the paper puts the 2 as a subscript. \\
%SnapKV \\
%Standardization Points:
%Bold best result and underline second best result in evaluation. \\
%Paragraph headings (that are bolded) should be consistent. Currently some end in periods, others do not and some end in colons. We should pick one and stick to it throughout.}

\section*{Acknowledgments}

We sincerely appreciate the anonymous reviewers. Their insightful feedback helps significantly improve the quality of the paper. This research was supported by the National Science Foundation (NSF) under Grant No. 2441601. The work utilized the Delta and DeltaAI system at the National Center for Supercomputing Applications (NCSA) and Jetstream2 at Indiana University through allocation CIS240055 from the Advanced Cyberinfrastructure Coordination Ecosystem: Services \& Support (ACCESS) program, which is supported by National Science Foundation grants \#2138259, \#2138286, \#2138307, \#2137603, and \#2138296. The Delta advanced computing resource is a collaborative effort between the University of Illinois Urbana-Champaign and NCSA, supported by the NSF (award OAC 2005572) and the State of Illinois. UIUC SSAIL Lab is supported by research funding and gift from IBM, Amazon, Google, and AMD. 

\bibliography{reference}
\bibliographystyle{colm2026/colm2026_conference}

\appendix
\newpage
\section{Training, Model, and Evaluation Details}\label{app:experiment_setup}
\paragraph{Training:}All models are trained across 16 NVIDIA A100s in parallel. We employ a learning rate of $3e-4$, with warmup over 2k steps and cosine decay. Optimizer is AdamW with $\lambda = (.9, .95)$ and weight decay $0.1$. 

The models presented in Section~\ref{sec:main_results}, \ref{sec:cle}, and \ref{sec:empirical} are trained on 50B tokens, while the models for ablation in Section~\ref{sec:ablations} are trained on 10B tokens. The training data is drawn from Dolma v1.7~\citep{dolma} using the data-mixture reweighting adopted by~\citet{bamba}, with a context length of 4096 and 0.5M tokens per minibatch.

\paragraph{Model:}
% We utilize the standard Llama3 tokenizer~\cite{llama3} for all models, with a vocabulary size of 128256. All models are 32 layers, with $d_{\mathrm{model}}=1280$. 
% \begin{itemize}
%     \item Universal Attention \& Llama3: $n_{\mathrm{heads}}=16$, $n_{\mathrm{kv}}=4$, $d_{\mathrm{ff}}=4096$, and $\theta_{\mathrm{RoPE}}=5\times 10^5$.
%     \item Mamba2: Default values from mamba\_ssm. 
%     \item Gated Delta Net: $n_{\mathrm{heads}}=4$, $d_{\mathrm{head}}=256$, $d_{\mathrm{ff}}=3072$. 
%     \item SnapKV: $r_\text{sparsity}=0.1$, $\text{window\_size}=32$, $\text{kernel\_size}=7$, maxpooling.
%     \item H$_2$O: $r_\text{heavy}=0.05$, $r_\text{recent}=0.05$.
% \end{itemize}
% For context extension on Llama3 and Universal Attention, we use YaRN with $(\beta_{\mathrm{slow}}, \beta_{\mathrm{fast}})=(0.7, 35)$. 
% The reduction on $d_{\mathrm{ff}}$ for Mamba2 and GDN is to balance the additional parameters introduced by Gated Delta Net, and keep the total number of parameters at $\sim1\text{B}$.   
% For the hyrbid models, we swap the 4, 12, 20, 28 layers using a GQA with $n_{\mathrm{heads}}=16$, $n_{\mathrm{kv}}=4$, $d_{\mathrm{ff}}=3072$. For GDN-H, we keep $\theta_{\mathrm{RoPE}}=5\times 10^5$. For Mamba2-H, we remove RoPE.
We use the Llama-3 tokenizer~\citep{llama3} (vocabulary size 128{,}256)
for all models. Unless noted below, all models have 32 layers with
$d_{\mathrm{model}}=1280$.

\begin{itemize}
    \item Base architectures:

\begin{itemize}
    \item \textbf{Universal Attention / Llama-3:}
      $n_{\mathrm{heads}}{=}16$, $n_{\mathrm{kv}}{=}4$,
      $d_{\mathrm{ff}}{=}4096$, $\theta_{\mathrm{RoPE}}{=}5\!\times\!10^5$.
      Context extension uses YaRN with
      $(\beta_{\mathrm{slow}},\beta_{\mathrm{fast}})=(0.7, 35)$.
    \item \textbf{Mamba2:}
      $d_{\mathrm{state}}{=}128$, $d_{\mathrm{conv}}{=}4$,
      $d_{\mathrm{ff}}{=}3072$.
    \item \textbf{Gated DeltaNet:}
      $n_{\mathrm{heads}}{=}4$, $d_{\mathrm{head}}{=}256$,
      $d_{\mathrm{ff}}{=}3072$.
\end{itemize}
For Mamba2 and Gated DeltaNet, $d_{\mathrm{ff}}$ is reduced from 4096 to keep total size ${\approx}1$B.

\item Hybrid variants (GDN-H, Mamba2-H):
In both hybrids, layers 4, 12, 20, and 28 are replaced with
grouped-query attention ($n_{\mathrm{heads}}{=}16$,
$n_{\mathrm{kv}}{=}4$). %, 
% $d_{\mathrm{ff}}{=}3072$).
GDN-H retains $\theta_{\mathrm{RoPE}}{=}5\!\times\!10^5$;
Mamba2-H uses no positional encoding in its attention layers following~\citet{nemotron-h}.

\item KV-cache compression baselines:
\begin{itemize}
    \item \textbf{SnapKV:} sparsity ratio $r{=}0.1$,
      \texttt{window\_size}$\,{=}\,$32,
      \texttt{kernel\_size}$\,{=}\,$7, max pooling.
    \item \textbf{H$_2$O:} $r_{\mathrm{heavy}}{=}0.05$,
      $r_{\mathrm{recent}}{=}0.05$.
\end{itemize}

\end{itemize}
\paragraph{Evaluation:}We utilize \texttt{EleutherAI\textbackslash lm-evaluation-harness}~\citep{lm-eval-harness} for evaluating on LongBench-E~\citep{longbench} and the common-sense reasoning benchmarks: WikiText~\citep{wikitext}, LAMBADA~\citep{lambada}, PIQA~\citep{piqa}, HellaSwag~\citep{hellaswag}, WinoGrande~\citep{winogrande}, ARC-Easy and ARC-Challenge~\citep{arc}, Social IQa~\citep{siqa}, and BoolQ~\citep{boolq}.
For RULER~\citep{ruler}, we utilize the official GitHub repository\footnote{\url{https://github.com/NVIDIA/RULER}}, following the default \texttt{synthetic} task set.

\section{Compute State Size \& KV-Cache Size}\label{app:kv_cache_comp}
Linear attention models are known for their constant state size, which does not grow as the input sequence length grows. However, at short sequence length, linear attention can often have larger state than a softmax attention with similar number of parameters. Here we present the steps we used to compute the state size in our Mamba2 model and Gated DeltaNet model, as well as the KV-cache size of the GQA model. 

\paragraph{Mamba2~\citep{mamba2}:} Following the official implementation of Mamba2\footnote{\url{https://github.com/state-spaces/mamba}}, we take the following parameters:
\begin{itemize}
    \item $d_\text{model}$
    \item $\texttt{expand}$, 2 by default, 
    \item $d_\text{state}$, 128 by default.
\end{itemize}
The state's shape is $(n_\text{heads}, d_\text{state}, d_\text{head})$, where the total number of elements is
\begin{align*}
    s_{\text{Mamba}} = &\ n_\text{heads}\times d_\text{state}\times d_\text{head}\\
    =&\ \frac{d_\text{model}\times\texttt{expand}}{d_\text{head}}\times d_\text{state}\times d_\text{head}\\
    =&\ \texttt{expand} \times d_\text{model}\times d_\text{state}.
\end{align*}

\paragraph{Gated DeltaNet~\citep{gated_delta_net}:} Following the official implementation listed in \texttt{flash-linear-attention}\footnote{\url{https://github.com/fla-org/flash-linear-attention}}, we take the following parameters:
\begin{itemize}
    \item $d_\text{head}$, 256 by default, 
    \item $\texttt{expand\_v}$, 2 by default, 
    \item $n_\text{heads}$, 6 by default.
\end{itemize}
The state's shape is $(\texttt{num\_heads}, \texttt{v\_head\_dim}, \texttt{head\_dim})$, where the total number of elements is
\begin{align*}
    s_{\text{GDN}} = &\ n_\text{heads}\times d_\text{head, v}\times d_\text{head}\\
    =&\ \texttt{expand\_v}\times n_\text{heads}\times d_\text{head}^2.
\end{align*}

\paragraph{Grouped-Query Attention~\citep{grouped_query_attention}:} Each entry of the GQA's KV cache contains a key and a value, where both of them share the size of $n_\text{heads, kv}\times d_\text{head}$. Thus, the total size of the KV-cache after seeing a sequence of length $l$ is
\begin{align*}
    s_{\text{GQA}} = 2\times n_\text{heads, kv}\times d_\text{head}\times l.
\end{align*}

To compare the actual memory footprint, we can unify them using the actual size of the stored states. Thus, for a sequence of length $l$, a hybrid model consisting of $n_\text{Mamba}$ Mamba2 layers, $n_\text{GDN}$ Gated DeltaNet layers, and $n_\text{GQA}$ grouped-query attention layers, has the following size of saved tensors:
\begin{align*}
    n_\text{Mamba}\times s_{\text{Mamba}} + n_\text{GDN}\times s_{\text{GDN}} + n_\text{GQA}\times s_{\text{GQA}}.
\end{align*}
Plugging-in the actual numbers from our setting as listed in Appendix~\ref{app:experiment_setup}, we got the following formula for the size of the stored states and KV-caches for each of our model configurations, as listed in Table~\ref{tab:state_size_math}. Here we store all states and KV-caches in bf16. 
\begin{table}[!ht]
\centering
\caption{Size of the stored states and KV-caches, w.r.t. sequence length $l$.}
\label{tab:state_size_math}
% \scriptsize
\begin{tabular}{lc}
\toprule
 \textbf{Model} &  \textbf{Stored Size} \\
\midrule
Mamba & $20\text{MB}$ \\
Mamba-Hybrid & $l\times5\text{KB} + 17.5\text{MB}$ \\
Gated-DeltaNet & $32\text{MB}$ \\
Gated-DeltaNet-Hybrid & $l\times5\text{KB} + 28\text{MB}$ \\
Llama3, Llama3-Gated & $l\times40\text{KB}$ \\
Llama3 with $10\times$ KV-cache eviction & $l\times4\text{KB}$ \\
\bottomrule
\end{tabular}
\end{table}

\newpage
\section{Kernel Design}
\label{app:kernel-design}

We detail a memory-efficient Universal Attention layer in listing~\ref{alg:fused-ua} alongside the triton kernels for both the forward and backwards passes in listings~\ref{alg:fused-ua-kern-fwd} \&~\ref{alg:fused-ua-kern-bwd}, respectively. 
%We detail our kernels in listings~\ref{alg:fused-ua-kern-fwd} \&~\ref{alg:fused-ua-kern-bwd}, demonstrating the triton kernels for both the forward and backwards passes, respectively. 

\textbf{Decay:} The decay mask kernels launch one kernel for both the forward and backwards pass. Thread-blocks tile over rows, computing the decay output (in the forward pass), or gradients (in the backwards pass) for a chunk of columns. 
%The decay-mask kernel fuses the three decay-mechanisms into one kernel. Thread-blocks tile over rows, computing the decay for a chunk of columns. The inner-loop in lines 16-23 tiles over the rows of the decay-mask and: (1) Loads two tiles of the queries (line 17), source-decays (line 18), and dest-decays (line 19), (2) computes \texttt{$\Gamma$c} - a local chunk of the decay mask - (line 20), (3) aggregates the local chunk into the running cumulative-sum, \texttt{osum}, and finally (4) writes to the decay-mask tensor in HBM (line 22).

\textbf{Memory-efficient attention:} The memory-efficient attention kernel launches one kernel in the forward pass and two in the backwards pass. In the forward pass, the attention kernel launches 2D thread-blocks with a block computing one chunk (consisting of multiple rows) of the output tensor. In the backwards pass, we first launch a preprocessing kernel that computes \texttt{o $\odot$ dO} similar to FlashAttention. Then we launch the main backwards kernel that computes the gradients of the queries keys and values. The two functions \texttt{mem\_efficient\_bwd\_dkv} and \texttt{mem\_efficient\_bwd\_dq} are two subroutines of the main kernel and follow logic similar to FlashAttention-2. However, prior to computing softmax normalized values, the decay mask is applied to intermediate attention outputs to correctly compute normalized attention scores. If GQA is applied, our kernel mimics a broadcast on the fly by indexing the correct kv-head of the decay-mask.

%The attention kernel launches 2D thread-blocks of size \texttt{[by, bx]}, with a block computing one chunk (consisting of multiple rows) of the output tensor \texttt{o}. A thread-block first loads queries (line 28) as \texttt{qi}, and iterates over tiles of keys and values (loop in lines 30-40). The loop first computes the dot-product of \texttt{qi} with a tile of keys (line 31), producing a tile of the attention matrix. Next, it indexes the decay-mask, extracting the correct head-dimension (mimicking a broadcast on the fly), to sum with the attention matrix. It then computes an online softmax (lines 33-38) updating metadata variables as required. The online softmax resembles flash-attention~\citep{}. Finally, it writes to the output tensor in HBM (line 41) after the loop concludes.

\definecolor{codered}{rgb}{0.6,0,0}
\definecolor{codegreen}{rgb}{0,0.6,0}
\definecolor{codeblue}{rgb}{0,0,0.6}
\definecolor{codepurple}{rgb}{0.58,0,0.82}
\definecolor{codebg}{rgb}{0.97,0.97,0.97}

\lstdefinestyle{fused-attn-style}{
    language=Python,
    frame=single,
    framerule=0pt,
    backgroundcolor=\color{codebg},
    commentstyle=\color{codepurple},
    keywordstyle=\color{codeblue},
    emphstyle=\ttfamily\color{codered},
    stringstyle=\color{codegreen},
    %basicstyle=\ttfamily\small,
    basicstyle=\ttfamily\linespread{0.85}\footnotesize,
    breakatwhitespace=false,
    breaklines=true,
    captionpos=b,
    keepspaces=true,
    numbers=left,
    numbersep=-4pt,
    showspaces=false,
    showstringspaces=false,
    showtabs=false,
    tabsize=2,
    morekeywords={},
    %emph={template, int, void, map, coord, Dim3, return}
}

%% Gives access to math mode in lstlistings. %%
\lstset{
  mathescape,
  literate={->}{$\rightarrow$}{2}
           {ε}{$\varepsilon$}{1}
}

\begin{lstlisting}[style={fused-attn-style},
    label={alg:fused-ua},
    caption={Pseudocode for a memory-efficient UA layer. A collection of kernels compute the decay-mask and attention output, additively applying the decay mask to the attention matrix in a fused online softmax kernel, avoiding expensive I/O to HBM. Activation-Checkpointing additionally avoids saving the quadratic decay mask per-layer.},
    float=h]
class UniversalAttention:
  # b (batch), nkv (kv-heads)
  # l (seq length), d (head-dim)
  def forward(qgate, kgate, Q, K, V):
    decay = decay_fwd(qgate, kgate, K)  # (b,nkv,l,l)
    o = mem_efficient_fwd(Q, K, V, decay)
    del decay
    return o

  def backward(dO):
    decay = decay_fwd(qgate, kgate, K)  # AC
    dq, dk, dv, ddecay = mem_efficient_bwd(Q, K, V, decay, dO)
    dqgate, dkgate, dkv2 = decay_bwd(ddecay)
    return dqgate, dkgate, dq, dk, dv
\end{lstlisting}

\definecolor{codered}{rgb}{0.6,0,0}
\definecolor{codegreen}{rgb}{0,0.6,0}
\definecolor{codeblue}{rgb}{0,0,0.6}
\definecolor{codepurple}{rgb}{0.58,0,0.82}
\definecolor{codebg}{rgb}{0.97,0.97,0.97}

\lstdefinestyle{fused-attn-style}{
    language=Python,
    frame=single,
    framerule=0pt,
    backgroundcolor=\color{codebg},   
    commentstyle=\color{codepurple},
    keywordstyle=\color{codeblue},
    emphstyle=\ttfamily\color{codered},
    stringstyle=\color{codegreen},
    %basicstyle=\ttfamily\small,
    basicstyle=\ttfamily\linespread{0.85}\footnotesize,
    breakatwhitespace=false,         
    breaklines=true,                 
    captionpos=b,                    
    keepspaces=true,                 
    numbers=left,                    
    numbersep=-4pt,                  
    showspaces=false,                
    showstringspaces=false,
    showtabs=false,                  
    tabsize=2,
    morekeywords={}
    %emph={template, int, void, map, coord, Dim3, return}
}

%% Gives access to math mode in lstlistings. %%
\lstset{
  mathescape,         
  literate={->}{$\rightarrow$}{2}
           {ε}{$\varepsilon$}{1}
}

\begin{lstlisting}[style={fused-attn-style}, 
    label={alg:fused-ua-kern-fwd},
    caption={Pseudocode for the decay forward and memory efficient forward kernels illustrating the per thread-block work.},
    float=t]
  # Assume block-size [by, bx].
  def decay_fwd(src, dest, K):
    b,n,f = tl.program_id(0), tl.program_id(1), tl.program_id(2)
    osum,srcsum,destsum = zeros(by,bx),zeros(by,bx),zeros(by,bx)
    for k in range(l//by+1):
      k, kt = K[b,n,k*by:k*(by+1),:], K[b,n,f*bx:k*(bx+1),:].T 
      srcsum += src[b,n,k*bx:k*(bx+1)] # broadcast and add
      destsum += dest[b,n,k*by:k*(by+1)] # broadcast and add
      $\Gamma$c = relu(dot(k,kt))**2 + srcsum + destsum # [by, bx]
      osum += log(1-$\Gamma$c)
      decay[b,n,k*by:k*(by+1),k*bx:k*(bx+1)] = osum 
    return decay
  # Omit batch dimension for brevity. 
  def memory_efficient_fwd(Q,K,V,decay):
    i,n = tl.program_id(0), tl.program_id(1)
    li, mi = [0]*by, [-inf]*by
    qi = Q[n,i*by:(i+1)*by, :]
    oi = zeros((by, d))
    for j in range(l//bx+1): 
      s = tl.matmul(qi, k[n%kvh,j*bx:(j+1)bx,:].T)+\   
            decay[n%kvh,i*by:(i+1)by,j*bx:(j+1)bx] # On-the-fly broadcast
      mij = tl.maximum(mi,tl.max(s,1))
      s -= mij
      p = tl.exp(s)
      lij = tl.sum(s, axis=-1)
      alpha = tl.exp(mi-mij)
      li, mi = li*alpha+lij, mij
      oi = oi * alpha + \
            tl.matmul(p,v[j*bx:(j+1)bx, :])
    store(oi)\end{lstlisting}

    \begin{lstlisting}[style={fused-attn-style}, 
    label={alg:fused-ua-kern-bwd},
    caption={Pseudocode for the decay backward and memory-efficient decay backward kernels illustrating the per thread-block work.},
    float=t]
  def decay_bwd(ddecay, src, dest, K):
    b,n,f = tl.program_id(0), tl.program_id(1), tl.program_id(2)
    dki, dsrci, suff = zeros(by,d), zeros(by), 0
    for j in rev_range(l//bx+1):
      ki = K[b,n,f*by:(f+1)*by,:]
      kj = K[b,n,j*bx:(j+1)*bx,:]
      $\Gamma$ = tl.dot(ki*sqrt(src), (kj*sqrt(dest)).T)
      $\Gamma$c = relu($\Gamma$)**(2/3)
      dd = rev_cumsum(ddecay[b,n,f*by:(f+1)*by,j*bx:(j+1)*bx]) + suff
      suff += tl.sum(dd, -1)
      dd = where($\Gamma$>0, -dd/(1-$\Gamma$c) * (2/3) * relu($\Gamma$)**(-1/3), 0)
      dki += tl.dot(dd, kj*sqrt(dest))
      dsrci += tl.sum(dd*$\Gamma$, -1) / (2*src)
    dk_i[b,n,f*by:(f+1)*by,:] = dki * sqrt(src)
    dsrc[b,n,f*by:(f+1)*by] = dsrci
    # Symmetric col-kernel for dk_j, ddest
    return dk_i+dk_j, dsrc, ddest
  # Preprocess: Di = rowsum(oi $\odot$ doi)
  # --- dK, dV kernel (key-parallel) ---
  def mem_efficient_bwd_dkv(Q,K,V,decay,dO):
    j,n = tl.program_id(0), tl.program_id(1)
    kj = K[n%kvh,j*bx:(j+1)*bx,:]
    vj = V[n%kvh,j*bx:(j+1)*bx,:]
    dkj, dvj = zeros(bx,d), zeros(bx,d)
    for i in range(l//by+1):
      qi = Q[n,i*by:(i+1)*by,:]
      aff = decay[n%kvh,i*by:(i+1)*by,j*bx:(j+1)*bx]
      s = tl.dot(kj, qi.T) + aff.T
      p = tl.exp(s - M[n,i*by:(i+1)*by])
      dvj += tl.dot(p, dO[n,i*by:(i+1)*by,:])
      dp = tl.dot(vj, dO[n,i*by:(i+1)*by,:].T)
      ds = p * (dp - D[n,i*by:(i+1)*by])
      dkj += tl.dot(ds, qi)
    DK[n%kvh,j*bx:(j+1)*bx,:] = dkj
    DV[n%kvh,j*bx:(j+1)*bx,:] = dvj
  # --- dQ, ddecay kernel (query-parallel) ---
  def mem_efficient_bwd_dq(Q,K,V,decay,dO):
    i,n = tl.program_id(0), tl.program_id(1)
    qi = Q[n,i*by:(i+1)*by,:]
    doi = dO[n,i*by:(i+1)*by,:]
    mi = M[n,i*by:(i+1)*by]
    dqi = zeros(by, d)
    for j in range(l//bx+1):
      kj = K[n%kvh,j*bx:(j+1)*bx,:]
      vj = V[n%kvh,j*bx:(j+1)*bx,:]
      aff = decay[n%kvh,i*by:(i+1)*by,j*bx:(j+1)*bx]
      s = tl.dot(qi, kj.T) + aff
      p = tl.exp(s - mi)
      dp = tl.dot(doi, vj.T)
      ds = p * (dp - D[n,i*by:(i+1)*by])
      ddecay[n%kvh,i*by:(i+1)*by,j*bx:(j+1)*bx] = ds
      dqi += tl.dot(ds, kj)
    DQ[n,i*by:(i+1)*by,:] = dqi\end{lstlisting}

\newpage
\color{edits}
% \section{Rebuttal Results}

\section{Hybrid Model Results}\label{app:hybrid_results}

We further evaluate UA in hybrid architectures that combine it with complementary sequence-modeling mechanisms. We consider two variants: \textbf{UA-Softmax}, which replaces one out of every eight UA layers with standard full attention, and \textbf{UA-Mamba}, which alternates UA and Mamba layers with an even split. All models follow the same training recipe as our main experiments. To isolate the architectural effects of hybridization, we evaluate these models without KV-cache pruning.

Table~\ref{tab:reb_hybrid_perf} shows that UA remains effective when incorporated into both hybrid architectures. UA-Softmax achieves performance close to pure UA, with RULER/LongBench-E averages of 48.8/17.6 compared with 48.5/18.1 for UA. UA-Mamba is slightly weaker overall at 47.9/15.9, but reduces the effective cache relative to the full-attention hybrid.

The hybrids also exhibit different capability trade-offs. In particular, introducing full-attention or Mamba layers substantially improves aggregation performance relative to pure UA, while reducing tracking performance. This suggests that UA can serve as a cache-efficient, high-resolution attention component within broader hybrid architectures, with full-attention or recurrent layers providing complementary memory behavior when stronger aggregation capability is desired.

\begin{table}[!ht]
\centering
\caption{Hybrid UA model performance, compared against pure Llama3 and UA models, with no eviction. 
% KV cache size reports the average token count retained across all layers.
}
\label{tab:reb_hybrid_perf}
\resizebox{0.9\textwidth}{!}{
\scriptsize
\begin{tabular}{lcccccc|cc}
\toprule
&\multicolumn{6}{c|}{\textbf{RULER (4k)}}& \multicolumn{2}{c}{\textbf{LongBench-E (0-4k)}}\\
\cmidrule(lr){2-7}\cmidrule(lr){8-9}
\textbf{Model} & \textbf{Retrieval} & \textbf{Tracking} & \textbf{Aggregation} & \textbf{QA} & \textbf{Avg.} & \textbf{Cache (MB)}  &\textbf{Avg.} & \textbf{Cache (MB)} \\
\midrule\midrule
\textbf{UA-Mamba} & 63.7 & 29.8 & 15.2 & 28.7 & 47.9 & 81.9 & 15.9 & 66.7\\
\textbf{UA-Softmax} & 64.3 & 33.5 & 11.7 & 31.5 & 48.8 & 143.7 & 17.6 & 113.3\\
\textbf{Llama3} & 54.2 & 8.4  & 13.6 & 30.4 & 40.8 & 143.7 & 16.4 & 113.3\\
\textbf{UA, p=0} & 65.5 & 37.0 & 5.1 & 29.3 & 48.5 & 143.7 & 18.1 & 113.3\\
\midrule\bottomrule
\end{tabular}
}
\end{table}

\section{Additional Ablation Results} \label{app:add_ablation}

We further evaluate the robustness of UA to alternative architectural choices and compare against individual decay mechanisms and sliding-window attention. Table~\ref{tab:reb_additional_ablation} reports averaged RULER \texttt{S1}, \texttt{S2}, and \texttt{S3} scores across context lengths for 1B models trained on 10B tokens.

UA remains effective when adopting Qwen3-style normalized queries and keys, with slightly improved performance at 16K. We also evaluate each decay mechanism in isolation. In particular, using only the query-aware decay $f^{(\text{q})}$ corresponds to the FoX-style design. While this variant performs reasonably at 4K, its performance degrades substantially with context length, reaching 40.0 at 16K compared with 58.4 for full UA. The key-aware and similarity-aware mechanisms likewise underperform the complete model when used alone, indicating that the three decay signals provide complementary retention behavior. Sliding-window attention is competitive only at the shortest context length and degrades sharply as context increases. Overall, these results show that UA is robust to alternative query/key normalization choices and that combining multiple decay mechanisms is important for robust long-context performance.

\begin{table}[!ht]
\centering
\caption{Additional ablations on model design choices. 1B models, trained on 10B tokens.}
%\caption{Impact of UA components on downstream retrieval performance}
\label{tab:reb_additional_ablation}
\begin{tabular}{lccc}
\toprule
 & \multicolumn{3}{c}{\textbf{RULER S1, S2, S3 (\% $\uparrow$)}} \\
\cmidrule(lr){2-4}
\textbf{Method} & \textbf{4k} & \textbf{8k} & \textbf{16k}\\
\midrule
\textbf{Universal Attention} & 99.6 & 94.2 & 58.4 \\
\quad Normalized QK (Qwen3-style) & 100.0 & 93.8 & 60.2 \\
\quad Only query-aware $f^{(\text{q})}$ (FoX) & 90.4 & 68.7 & 40.0 \\
\quad Only key-aware $f^{(\text{k})}$ & 96.8 & 65.3 & 54.8 \\
\quad Only similarity $f^{(\text{sim})}$ & 82.3 & 85.4 & 56.3 \\
Sliding-window attention ($w=512$) & 100.0 & 68.2 & 31.5 \\
\bottomrule
\end{tabular}
\end{table}%

\section{3B Model Results}\label{app:3B_model_results}
To evaluate whether UA's benefits persist at larger model scale, we train 3B-parameter Llama3-style softmax and UA models using the same 50B-token training recipe as in our main experiments. Table~\ref{tab:reb_3B_perf} reports RULER and LongBench-E results together with effective cache size.

At 3B scale, UA continues to improve both model quality and cache efficiency. On RULER at 4K context, UA achieves an average score of 52.6 compared with 48.8 for Llama3-3B, while $p{=}0.05$ reduces the effective cache from 341.8 MB to 46.6 MB, corresponding to a $7.3\times$ reduction. On LongBench-E (0--4K), UA reaches 23.5 compared with 21.5 for Llama3-3B while reducing cache from 271.9 MB to 41.5 MB, a $6.5\times$ reduction. These results indicate that UA's quality--memory advantages extend beyond the 1B setting.

\begin{table}[!ht]
\centering
\caption{Llama3-3B and UA-3B model performance, trained on 50B tokens. 
% KV cache size reports the average token count retained across all layers.
}
\label{tab:reb_3B_perf}
\resizebox{0.9\textwidth}{!}{
\scriptsize
\begin{tabular}{lcccccc|cc}
\toprule
&\multicolumn{6}{c|}{\textbf{RULER (4k)}}& \multicolumn{2}{c}{\textbf{LongBench-E (0-4k)}}\\
\cmidrule(lr){2-7}\cmidrule(lr){8-9}
\textbf{Model} & \textbf{Retrieval} & \textbf{Tracking} & \textbf{Aggregation} & \textbf{QA} & \textbf{Avg.}  & \textbf{Cache (MB)} &\textbf{Avg.} & \textbf{Cache (MB)}\\
\midrule\midrule
Llama3-3B & 64.7 & 22.8  & 16.6 & 30.1 & 48.8 & 341.8 & 21.5 & 271.9\\
UA-3B, p=0 & 69.8 & 40.3 & 7.2 & 35.4 & 52.6 & 341.8 & 23.6 & 271.9\\
UA-3B, p=0.05 & 70.3 & 38.7 & 6.8 & 34.4 & 52.6 & 46.6 & 23.5 & 41.5\\
\midrule\bottomrule
\end{tabular}
}
\end{table}

\section{Post-Training Adaptation}\label{app:post_train_results}
Although UA is designed as a distinct attention architecture and is primarily evaluated through pretraining from scratch, we also investigate whether it can be introduced into an existing pretrained Transformer checkpoint through continued training.

Starting from our pretrained Llama3 model, we convert its attention layers to UA while retaining the pretrained Llama3 weights, randomly initialize the UA-specific parameters---including the low-rank attention-gate projections and decay parameters---and continue training for an additional 10B tokens. Table~\ref{tab:reb_post_train_llama3} reports the resulting RULER and LongBench-E performance together with effective cache size.

\begin{table}[!ht]
\centering
\caption{Post-training adaptation of a pretrained Llama3-1B checkpoint to UA. 
% KV cache size reports the average token count retained across all layers.
}
\label{tab:reb_post_train_llama3}
\resizebox{0.9\textwidth}{!}{
\scriptsize
\begin{tabular}{lcccccc|cc}
\toprule
&\multicolumn{6}{c|}{\textbf{RULER (4k)}}& \multicolumn{2}{c}{\textbf{LongBench-E (0-4k)}}\\
\cmidrule(lr){2-7}\cmidrule(lr){8-9}
\textbf{Model} & \textbf{Retrieval} & \textbf{Tracking} & \textbf{Aggregation} & \textbf{QA} & \textbf{Avg.}  & \textbf{Cache (MB)} &\textbf{Avg.} & \textbf{Cache (MB)} \\
\midrule\midrule
Llama3 & 54.2 & 8.4 & 13.6 & 30.4 & 40.8 & 143.7 & 16.4 & 113.3\\
Llama3-UA, p=0 & 58.7 & 16.2 & 10.8 & 32.4 & 44.0 & 143.7 & 17.1 & 113.3\\
Llama3-UA, p=0.05 & 59.3 & 16.2 & 9.7 & 33.0 & 44.3 & 38.6 & 17.2 & 41.3\\
\midrule
UA, p=0 & 65.5 & 37.0 & 5.1 & 29.3 & 48.5 & 143.7 & 18.1 & 113.3\\
UA, p=0.05 & 68.4 & 36.4 & 5.7 & 29.5 & 50.3 & 21.4 & 17.9 & 20.2\\
\midrule\bottomrule
\end{tabular}
}
\end{table}
Post-training adaptation substantially improves over the original Llama3 baseline while enabling significant cache compression. At $p{=}0.05$, Llama3-UA improves the RULER average from 40.8 to 44.3 while reducing the cache from 142.4 MB to 38.6 MB; on LongBench-E, it improves from 16.4 to 17.2 while reducing the cache from 113.3 MB to 41.3 MB. The adapted model remains below UA trained from scratch, which reaches a RULER average of 50.3 with a 21.4 MB cache at the same pruning threshold. These results suggest that UA can be incorporated into existing checkpoints through continued training, although training from scratch currently yields the strongest performance--memory trade-off.

\section{Latency Analysis}
\label{app:latency}
We evaluate the end-to-end inference-time efficiency of UA. An inference-optimized dynamic cache manager is not implemented in the current system, as efficiently operating on dynamically pruned KV caches requires sparse matrix--vector computation and specialized sparse kernels, which remain active areas of systems research~\citep{asyncsparse, sparse_kernel}. Nevertheless, the current implementation already provides measurable benefits during training and decoding.

For prefill, our custom UA kernels improve end-to-end training throughput by approximately 1.13$\times$; since prefill exercises the same forward attention path, this suggests that kernel optimization can also reduce TTFT, although we do not report a matched inference-prefill baseline here. For decoding, we evaluate a naive UA inference implementation with and without KV-cache pruning. As shown in Table~\ref{tab:latency_analysis}, pruning yields approximately 1.1$\times$--1.4$\times$ per-token decode speedup (11--29\% lower latency), with larger benefits at longer prompt lengths.

\begin{table}[!ht]
\centering
\caption{Prefill latency, per-step decode latency with and without KV-cache pruning, and post-eviction cache size at different sequence lengths.}
\label{tab:latency_analysis}
% \scriptsize
\begin{tabular}{l|c|cc|c}
\toprule
\textbf{Sequence }& \textbf{Prefill} & \multicolumn{2}{c|}{\textbf{Decode Latency per Step}}& \textbf{Cache Size} \\
\textbf{Length}	& \textbf{Latency}	& \textbf{w. full cache}	& \textbf{w. evicted cache}	& \textbf{after eviction (MB)}\\
\midrule
3,678	& 438 ms & 1.1 ms & 978 µs & 21.5\\
7,875 & 1.76 s & 1.2 ms & 998 µs & 24.4\\
15,678 & 5.44 s & 1.4 ms & 997 µs & 24.3\\
\bottomrule
\end{tabular}
\end{table}

\section{Detailed RULER results for Figure~\ref{fig:RULER_LongBench}}\label{app:ruler_16k}
We report the complete RULER results underlying Figure~\ref{fig:RULER_LongBench}, including per-task scores for all models at context lengths from 1K to 16K and their corresponding effective cache sizes. In addition to the configurations shown in the main comparison, we include GDN-H with YaRN at 8K and 16K for a consistent long-context comparison, as well as full-cache UA ($p{=}0$) to separate the effects of the UA architecture from those of cache pruning. Applying YaRN substantially improves GDN-H at long context, particularly on retrieval, but UA with $p{=}0.05$ still achieves a better overall RULER score with a smaller effective cache at both 8K and 16K.

\begin{table}[!ht]
\centering
\caption{Detailed RULER (1k) scores, with effective cache size. }
\label{tab:reb_1k_ruler}
\resizebox{0.95\textwidth}{!}{
\scriptsize
\begin{tabular}{lcccccccccccccccc}
\toprule
 % & \multicolumn{2}{c|}{\textbf{Perplexity ($\downarrow$)}} & \multicolumn{10}{c}{\textbf{Zero-Shot Common-Sense Reasoning (acc $\uparrow$)}} \\
% \cmidrule(lr){2-3} \cmidrule(lr){4-13}
\textbf{Model} & \textbf{S1} & \textbf{S2} & \textbf{S3} & \textbf{M1} & \textbf{M2} & \textbf{M3} & \textbf{MV} & \textbf{MQ} & \textbf{VT} & \textbf{CWE} & \textbf{FWE} & \textbf{QA1} & \textbf{QA2} & \textbf{Avg.} &\textbf{Cache (MB)} \\
\midrule\midrule
Mamba2 & 100.0 & 100.0 & 100.0 & 47.2 & 4.4 & 4.2 & 53.0 & 54.5 & 1.8 & 3.6 & 43.1 & 44.0 & 5.8 & 43.2 & 20.0\\
Mamba2-H & 100.0 & 100.0 & 96.8 & 87.0 & 68.0 & 37.6 & 72.4 & 73.0 & 5.4 & 20.0 & 43.1 & 44.4 & 5.0 & 57.9 & 21.5\\
GDN & 100.0 & 100.0 & 99.8 & 38.6 & 6.4 & 1.4 & 80.1 & 81.0 & 5.4 & 54.3 & 17.7 & 43.8 & 6.8 & 48.9 & 32.0\\
GDN-H & 100.0 & 100.0 & 99.2 & 94.0 & 97.4 & 43.2 & 83.0 & 74.8 & 38.5 & 56.7 & 21.9 & 49.0 & 7.0 & 66.5 & 32.0\\
Llama3 & 100.0 & 100.0 & 96.6 & 89.6 & 41.2 & 23.8 & 83.6 & 80.5 & 8.3 & 17.4 & 35.1 & 49.4 & 5.8 & 56.2 & 32.2\\
\quad + SnapKV & 95.2 & 7.4 & 0.0 & 3.2 & 0.8 & 0.0 & 0.0 & 0.0 & 0.1 & 0.3 & 39.7 & 5.4 & 5.0 & 12.1 & 3.2\\
\quad + Pyramid SnapKV & 7.6 & 0.4 & 0.0 & 0.8 & 0.0 & 0.0 & 0.0 & 0.0 & 0.0 & 0.1 & 5.3 & 5.2 & 3.8 & 1.8 & 3.2\\
\quad + H$_2$O & 0.0 & 0.0 & 0.0 & 2.2 & 0.0 & 0.0 & 0.2 & 0.0 & 0.0 & 2.6 & 41.9 & 21.8 & 6.0 & 5.7 & 3.2\\
\quad + Pyramid H$_2$O & 0.0 & 0.0 & 0.0 & 2.2 & 0.0 & 0.0 & 0.2 & 0.0 & 0.0 & 2.5 & 38.5 & 18.4 & 5.0 & 5.1 & 3.2\\
Llama3-G  & 100.0 & 100.0 & 99.8 & 82.2 & 51.0 & 31.8 & 83.4 & 80.3 & 25.3 & 13.3 & 47.3 & 49.6 & 5.2 & 59.2 & 32.2\\
\cellcolor{customblue} \textbf{UA, p=0} & \cellcolor{customblue} 100.0 & \cellcolor{customblue} 100.0 & \cellcolor{customblue} 100.0 & \cellcolor{customblue} 93.6 & \cellcolor{customblue} 90.4 & \cellcolor{customblue} 74.6 & \cellcolor{customblue} 90.6 & \cellcolor{customblue} 80.2 & \cellcolor{customblue} 53.0 & \cellcolor{customblue} 11.6 & \cellcolor{customblue} 22.3 & \cellcolor{customblue} 48.8 & \cellcolor{customblue} 5.2 & \cellcolor{customblue} 66.9 & \cellcolor{customblue} 32.2\\
\cellcolor{customblue} \textbf{UA, p=0.05} & \cellcolor{customblue} 100.0 & \cellcolor{customblue} 100.0 & \cellcolor{customblue} 99.8 & \cellcolor{customblue} 89.0 & \cellcolor{customblue} 92.2 & \cellcolor{customblue} 71.2 & \cellcolor{customblue} 87.0 & \cellcolor{customblue} 80.4 & \cellcolor{customblue} 53.0 & \cellcolor{customblue} 11.6 & \cellcolor{customblue} 23.3 & \cellcolor{customblue} 46.2 & \cellcolor{customblue} 5.6 & \cellcolor{customblue} 66.1 & \cellcolor{customblue} 10.2\\
\cellcolor{customblue} \textbf{UA, p=0.15} & \cellcolor{customblue} 100.0 & \cellcolor{customblue} 100.0 & \cellcolor{customblue} 100.0 & \cellcolor{customblue} 83.2 & \cellcolor{customblue} 86.0 & \cellcolor{customblue} 63.2 & \cellcolor{customblue} 79.5 & \cellcolor{customblue} 74.6 & \cellcolor{customblue} 53.8 & \cellcolor{customblue} 11.6 & \cellcolor{customblue} 17.5 & \cellcolor{customblue} 47.2 & \cellcolor{customblue} 5.4 & \cellcolor{customblue} 63.2 & \cellcolor{customblue} 7.2\\
\midrule\bottomrule
\end{tabular}
}
\end{table}

\begin{table}[!ht]
\centering
\caption{Detailed RULER (2k) scores, with effective cache size. }
\label{tab:reb_2k_ruler}
\resizebox{0.95\textwidth}{!}{
\scriptsize
\begin{tabular}{lcccccccccccccccc}
\toprule
 % & \multicolumn{2}{c|}{\textbf{Perplexity ($\downarrow$)}} & \multicolumn{10}{c}{\textbf{Zero-Shot Common-Sense Reasoning (acc $\uparrow$)}} \\
% \cmidrule(lr){2-3} \cmidrule(lr){4-13}
\textbf{Model} & \textbf{S1} & \textbf{S2} & \textbf{S3} & \textbf{M1} & \textbf{M2} & \textbf{M3} & \textbf{MV} & \textbf{MQ} & \textbf{VT} & \textbf{CWE} & \textbf{FWE} & \textbf{QA1} & \textbf{QA2} & \textbf{Avg.} &\textbf{Cache (MB)} \\
\midrule\midrule
Mamba2 & 100.0 & 96.0 & 98.6 & 32.8 & 1.2 & 0.0 & 29.8 & 29.9 & 0.9 & 1.3 & 30.7 & 22.2 & 23.0 & 35.9 & 20.0\\
Mamba2-H & 100.0 & 100.0 & 91.6 & 84.4 & 38.2 & 27.4 & 64.5 & 63.3 & 2.1 & 6.9 & 39.3 & 36.8 & 21.0 & 52.0 & 25.8\\
GDN & 100.0 & 100.0 & 98.8 & 34.6 & 2.0 & 0.2 & 58.2 & 65.7 & 4.5 & 9.8 & 22.1 & 17.6 & 23.6 & 41.3 & 32.0\\
GDN-H & 100.0 & 100.0 & 99.8 & 92.2 & 85.8 & 31.8 & 70.6 & 66.5 & 6.9 & 5.4 & 10.3 & 26.8 & 27.4 & 55.7 & 36.3\\
Llama3 & 100.0 & 100.0 & 94.0 & 82.2 & 16.4 & 9.6 & 72.9 & 69.5 & 3.1 & 4.7 & 30.6 & 35.2 & 24.4 & 49.4 & 66.3\\
\quad + SnapKV & 100.0 & 29.0 & 0.0 & 18.8 & 0.6 & 0.0 & 0.2 & 0.1 & 6.1 & 1.9 & 27.1 & 24.0 & 20.4 & 17.6 & 6.6\\
\quad + Pyramid SnapKV & 100.0 & 28.8 & 0.0 & 16.0 & 0.4 & 0.0 & 0.0 & 0.0 & 12.1 & 0.6 & 19.3 & 15.4 & 19.0 & 16.3 & 6.6\\
\quad + H$_2$O & 2.4 & 2.4 & 2.4 & 3.0 & 2.4 & 0.0 & 0.5 & 0.5 & 0.0 & 2.7 & 24.0 & 15.6 & 21.0 & 5.9 & 6.6\\
\quad + Pyramid H$_2$O & 2.4 & 2.4 & 2.4 & 3.0 & 2.2 & 0.0 & 0.7 & 0.9 & 0.1 & 2.9 & 23.5 & 15.6 & 20.6 & 5.9 & 6.6\\
Llama3-G  & 100.0 & 100.0 & 99.8 & 73.8 & 19.6 & 14.0 & 60.6 & 64.0 & 12.9 & 5.3 & 44.1 & 34.8 & 26.0 & 50.4 & 66.3\\
\cellcolor{customblue} \textbf{UA, p=0} & \cellcolor{customblue} 100.0 & \cellcolor{customblue} 100.0 & \cellcolor{customblue} 99.8 & \cellcolor{customblue} 63.6 & \cellcolor{customblue} 72.6 & \cellcolor{customblue} 68.5 & \cellcolor{customblue} 80.2 & \cellcolor{customblue} 63.5 & \cellcolor{customblue} 36.1 & \cellcolor{customblue} 5.7 & \cellcolor{customblue} 13.2 & \cellcolor{customblue} 37.8 & \cellcolor{customblue} 28.4 & \cellcolor{customblue} 59.2 & \cellcolor{customblue} 66.3\\
\cellcolor{customblue} \textbf{UA, p=0.05} & \cellcolor{customblue} 100.0 & \cellcolor{customblue} 100.0 & \cellcolor{customblue} 99.4 & \cellcolor{customblue} 62.4 & \cellcolor{customblue} 68.0 & \cellcolor{customblue} 63.2 & \cellcolor{customblue} 77.8 & \cellcolor{customblue} 67.6 & \cellcolor{customblue} 38.1 & \cellcolor{customblue} 6.3 & \cellcolor{customblue} 7.6 & \cellcolor{customblue} 37.4 & \cellcolor{customblue} 27.0 & \cellcolor{customblue} 58.1 & \cellcolor{customblue} 14.7\\
\cellcolor{customblue} \textbf{UA, p=0.15} & \cellcolor{customblue} 100.0 & \cellcolor{customblue} 100.0 & \cellcolor{customblue} 100.0 & \cellcolor{customblue} 60.4 & \cellcolor{customblue} 60.8 & \cellcolor{customblue} 55.0 & \cellcolor{customblue} 74.9 & \cellcolor{customblue} 65.7 & \cellcolor{customblue} 41.8 & \cellcolor{customblue} 5.5 & \cellcolor{customblue} 2.3 & \cellcolor{customblue} 37.8 & \cellcolor{customblue} 26.4 & \cellcolor{customblue} 56.2 & \cellcolor{customblue} 9.6\\
\midrule\bottomrule
\end{tabular}
}
\end{table}

\begin{table}[!ht]
\centering
\caption{Detailed RULER (4k) scores, with effective cache size. }
\label{tab:reb_4k_ruler}
\resizebox{0.95\textwidth}{!}{
\scriptsize
\begin{tabular}{lcccccccccccccccc}
\toprule
 % & \multicolumn{2}{c|}{\textbf{Perplexity ($\downarrow$)}} & \multicolumn{10}{c}{\textbf{Zero-Shot Common-Sense Reasoning (acc $\uparrow$)}} \\
% \cmidrule(lr){2-3} \cmidrule(lr){4-13}
\textbf{Model} & \textbf{S1} & \textbf{S2} & \textbf{S3} & \textbf{M1} & \textbf{M2} & \textbf{M3} & \textbf{MV} & \textbf{MQ} & \textbf{VT} & \textbf{CWE} & \textbf{FWE} & \textbf{QA1} & \textbf{QA2} & \textbf{Avg.} &\textbf{Cache (MB)} \\
\midrule\midrule
Mamba2 & 100.0 & 34.4 & 49.2 & 24.4 & 0.2 & 0.0 & 20.5 & 19.3 & 0.8 & 3.0 & 33.3 & 22.2 & 17.8 & 25.0 & 20.0\\
Mamba2-H & 100.0 & 100.0 & 60.4 & 69.6 & 2.4 & 9.2 & 58.8 & 57.3 & 3.2 & 18.4 & 23.3 & 34.2 & 25.2 & 43.2 & 35.5\\
GDN & 100.0 & 65.2 & 56.2 & 29.2 & 0.2 & 0.0 & 25.7 & 25.3 & 2.6 & 14.4 & 21.9 & 17.4 & 19.4 & 29.0 & 32.0\\
GDN-H & 100.0 & 100.0 & 91.2 & 74.8 & 25.0 & 17.6 & 34.8 & 36.0 & 4.4 & 36.3 & 2.9 & 27.0 & 22.8 & 44.1 & 46.0\\
Llama3 & 100.0 & 99.8 & 87.2 & 66.8 & 4.8 & 6.4 & 31.0 & 37.5 & 8.4 & 7.3 & 19.9 & 34.6 & 26.2 & 40.8 & 143.7\\
\quad + SnapKV & 99.8 & 19.4 & 0.0 & 38.8 & 0.4 & 0.0 & 10.0 & 5.0 & 13.4 & 1.1 & 13.9 & 29.4 & 21.8 & 19.5 & 14.4\\
\quad + Pyramid SnapKV & 100.0 & 23.2 & 0.2 & 33.2 & 0.2 & 0.0 & 6.3 & 6.3 & 20.3 & 0.6 & 11.9 & 28.0 & 22.2 & 19.4 & 14.4\\
\quad + H$_2$O & 4.6 & 6.2 & 5.6 & 7.2 & 1.2 & 0.0 & 5.3 & 3.1 & 0.2 & 4.8 & 16.5 & 18.6 & 21.2 & 7.3 & 14.4\\
\quad + Pyramid H$_2$O & 4.6 & 6.2 & 5.6 & 7.2 & 1.2 & 0.0 & 5.9 & 3.7 & 0.2 & 4.0 & 15.3 & 18.2 & 21.8 & 7.2 & 14.4\\
Llama3-G  & 100.0 & 100.0 & 97.6 & 61.2 & 2.4 & 7.2 & 26.5 & 34.7 & 16.3 & 9.8 & 33.0 & 33.4 & 23.2 & 41.9 & 143.7\\
\cellcolor{customblue} \textbf{UA, p=0} & \cellcolor{customblue} 100.0 & \cellcolor{customblue} 100.0 & \cellcolor{customblue} 82.8 & \cellcolor{customblue} 64.4 & \cellcolor{customblue} 77.6 & \cellcolor{customblue} 54.0 & \cellcolor{customblue} 30.0 & \cellcolor{customblue} 15.4 & \cellcolor{customblue} 37.0 & \cellcolor{customblue} 10.0 & \cellcolor{customblue} 0.2 & \cellcolor{customblue} 35.8 & \cellcolor{customblue} 22.8 & \cellcolor{customblue} 48.5 & \cellcolor{customblue} 143.7\\
\cellcolor{customblue} \textbf{UA, p=0.05} & \cellcolor{customblue} 100.0 & \cellcolor{customblue} 100.0 & \cellcolor{customblue} 86.2 & \cellcolor{customblue} 73.6 & \cellcolor{customblue} 73.8 & \cellcolor{customblue} 57.0 & \cellcolor{customblue} 34.7 & \cellcolor{customblue} 22.2 & \cellcolor{customblue} 36.4 & \cellcolor{customblue} 11.4 & \cellcolor{customblue} 0.1 & \cellcolor{customblue} 36.2 & \cellcolor{customblue} 22.8 & \cellcolor{customblue} 50.3 & \cellcolor{customblue} 21.4\\
\cellcolor{customblue} \textbf{UA, p=0.15} & \cellcolor{customblue} 100.0 & \cellcolor{customblue} 100.0 & \cellcolor{customblue} 86.6 & \cellcolor{customblue} 80.2 & \cellcolor{customblue} 66.6 & \cellcolor{customblue} 46.0 & \cellcolor{customblue} 33.9 & \cellcolor{customblue} 25.2 & \cellcolor{customblue} 30.6 & \cellcolor{customblue} 13.8 & \cellcolor{customblue} 0.0 & \cellcolor{customblue} 36.6 & \cellcolor{customblue} 21.6 & \cellcolor{customblue} 49.3 & \cellcolor{customblue} 12.9\\
\midrule\bottomrule
\end{tabular}
}
\end{table}

\begin{table}[!ht]
\centering
\caption{Detailed RULER (8k) scores, with effective cache size. }
\label{tab:reb_8k_ruler}
\resizebox{0.95\textwidth}{!}{
\scriptsize
\begin{tabular}{lcccccccccccccccc}
\toprule
 % & \multicolumn{2}{c|}{\textbf{Perplexity ($\downarrow$)}} & \multicolumn{10}{c}{\textbf{Zero-Shot Common-Sense Reasoning (acc $\uparrow$)}} \\
% \cmidrule(lr){2-3} \cmidrule(lr){4-13}
\textbf{Model} & \textbf{S1} & \textbf{S2} & \textbf{S3} & \textbf{M1} & \textbf{M2} & \textbf{M3} & \textbf{MV} & \textbf{MQ} & \textbf{VT} & \textbf{CWE} & \textbf{FWE} & \textbf{QA1} & \textbf{QA2} & \textbf{Avg.} &\textbf{Cache (MB)} \\
\midrule\midrule
Mamba2 & 100.0 & 24.8 & 24.0 & 12.8 & 0.0 & 0.0 & 7.4 & 6.6 & 1.6 & 2.2 & 29.8 & 11.2 & 15.2 & 18.1 & 20.0\\
Mamba2-H & 0.0 & 0.0 & 0.0 & 0.0 & 0.0 & 1.4 & 0.0 & 0.0 & 0.0 & 0.0 & 3.5 & 3.6 & 1.8 & 0.8 & 54.9\\
GDN & 100.0 & 47.4 & 30.8 & 27.0 & 0.0 & 0.0 & 23.3 & 23.0 & 2.7 & 1.3 & 21.7 & 9.8 & 15.2 & 23.2 & 32.0\\
GDN-H & 56.6 & 52.0 & 36.8 & 45.2 & 44.8 & 1.6 & 24.9 & 24.6 & 17.2 & 1.2 & 0.1 & 12.2 & 19.2 & 25.9 & 65.4\\
\quad + YaRN & 95.4 & 72.4 & 40.0 & 30.2 & 44.8 & 1.0 & 28.7 & 14.8 & 15.3 & 1.0 & 0.0 & 16.6 & 18.5 & 29.1 & 65.4\\
Llama3 & 99.6 & 95.0 & 73.0 & 51.2 & 2.2 & 1.2 & 24.7 & 27.6 & 2.1 & 5.2 & 13.7 & 15.6 & 18.4 & 33.0 & 299.0\\
\quad + SnapKV & 99.6 & 16.2 & 1.2 & 31.6 & 1.4 & 0.0 & 10.6 & 4.0 & 3.7 & 1.4 & 7.4 & 12.8 & 19.2 & 16.1 & 29.9\\
\quad + Pyramid SnapKV & 99.8 & 21.2 & 1.4 & 29.8 & 0.8 & 0.0 & 12.3 & 5.5 & 3.4 & 0.8 & 6.4 & 12.8 & 20.0 & 16.5 & 29.9\\
\quad + H$_2$O & 2.8 & 3.8 & 4.8 & 5.4 & 0.4 & 0.0 & 2.9 & 2.7 & 0.0 & 3.9 & 10.0 & 12.4 & 18.6 & 5.2 & 29.9\\
\quad + Pyramid H$_2$O & 2.8 & 3.8 & 4.8 & 5.4 & 0.4 & 0.0 & 3.4 & 3.1 & 0.0 & 3.3 & 9.0 & 10.4 & 18.4 & 5.0 & 29.9\\
Llama3-G  & 96.8 & 86.6 & 77.6 & 46.0 & 1.2 & 1.4 & 17.2 & 22.9 & 4.1 & 4.8 & 27.9 & 14.0 & 18.8 & 32.2 & 299.0\\
\cellcolor{customblue} \textbf{UA, p=0} & \cellcolor{customblue} 99.6 & \cellcolor{customblue} 78.0 & \cellcolor{customblue} 65.2 & \cellcolor{customblue} 49.0 & \cellcolor{customblue} 42.8 & \cellcolor{customblue} 14.6 & \cellcolor{customblue} 20.2 & \cellcolor{customblue} 7.5 & \cellcolor{customblue} 32.2 & \cellcolor{customblue} 6.1 & \cellcolor{customblue} 0.0 & \cellcolor{customblue} 13.0 & \cellcolor{customblue} 20.0 & \cellcolor{customblue} 34.5 & \cellcolor{customblue} 299.0\\
\cellcolor{customblue} \textbf{UA, p=0.05} & \cellcolor{customblue} 99.4 & \cellcolor{customblue} 75.2 & \cellcolor{customblue} 64.2 & \cellcolor{customblue} 46.8 & \cellcolor{customblue} 47.2 & \cellcolor{customblue} 12.2 & \cellcolor{customblue} 19.6 & \cellcolor{customblue} 9.0 & \cellcolor{customblue} 32.6 & \cellcolor{customblue} 6.0 & \cellcolor{customblue} 0.0 & \cellcolor{customblue} 12.0 & \cellcolor{customblue} 19.0 & \cellcolor{customblue} 34.1 & \cellcolor{customblue} 22.7\\
\cellcolor{customblue} \textbf{UA, p=0.15} & \cellcolor{customblue} 97.8 & \cellcolor{customblue} 53.0 & \cellcolor{customblue} 48.2 & \cellcolor{customblue} 40.4 & \cellcolor{customblue} 33.8 & \cellcolor{customblue} 9.2 & \cellcolor{customblue} 24.2 & \cellcolor{customblue} 11.1 & \cellcolor{customblue} 32.7 & \cellcolor{customblue} 2.4 & \cellcolor{customblue} 0.0 & \cellcolor{customblue} 10.4 & \cellcolor{customblue} 16.4 & \cellcolor{customblue} 29.2 & \cellcolor{customblue} 13.6\\
\midrule\bottomrule
\end{tabular}
}
\end{table}

\begin{table}[!ht]
\centering
\caption{Detailed RULER (16k) scores, with effective cache size. }
\label{tab:reb_16k_ruler}
\resizebox{0.95\textwidth}{!}{
\scriptsize
\begin{tabular}{lcccccccccccccccc}
\toprule
 % & \multicolumn{2}{c|}{\textbf{Perplexity ($\downarrow$)}} & \multicolumn{10}{c}{\textbf{Zero-Shot Common-Sense Reasoning (acc $\uparrow$)}} \\
% \cmidrule(lr){2-3} \cmidrule(lr){4-13}
\textbf{Model} & \textbf{S1} & \textbf{S2} & \textbf{S3} & \textbf{M1} & \textbf{M2} & \textbf{M3} & \textbf{MV} & \textbf{MQ} & \textbf{VT} & \textbf{CWE} & \textbf{FWE} & \textbf{QA1} & \textbf{QA2} & \textbf{Avg.} &\textbf{Cache (MB)} \\
\midrule\midrule
Mamba2            & 96.0 & 5.0 & 9.0 & 6.2 & 0.0 & 0.0 & 6.6 & 6.0 & 0.4 & 0.6 & 15.9 & 8.2 & 11.0 & 12.7 & 20.0\\
Mamba2-H     & 0.0 & 0.0 & 0.0 & 0.0  & 0.0 & 0.0 & 0.0 & 0.0 & 0.0 & 0.0 & 18.7 & 1.8 & 1.6 & 1.7 &  94.2\\
GDN    & 99.6 & 22.8 & 12.4 & 18.4 & 0.0 & 0.0 & 17.9 & 16.6 & 5.9 & 0.0    & 17.5 & 7.0 & 13.0 & 17.8 & 32.0 \\
GDN-H        & 28.4 & 20.2 & 14.6 & 20.6 & 8.0 & 0.8 & 11.5 & 9.6 & 0.0 & 0.1 & 0.2 & 10.6 & 16.0	& 10.8 & 104.7 \\
\quad + YaRN       & 92.6 & 27.2 & 25.4 & 18.9 & 9.2 & 1.0 & 15.4 & 6.2 & 0.0 & 0.4 & 0.0 & 14.3 & 16.4 & 17.5 & 104.7 \\
Llama3            & 99.4 & 68.6 & 35.8 & 26.4 & 0.2 & 0.0 & 15.3 & 7.1 & 1.6 & 3.8 & 11.3 & 10.4 & 17.0 & 22.8 &	 613.3\\
\quad + SnapKV & 99.8 & 6.0 & 0.2 & 10.8 & 0.4 & 0.0 & 5.3 & 1.1 & 2.6 & 2.2 & 7.0 & 11.6 & 17.4 & 12.6 & 61.3\\
\quad + Pyramid SnapKV & 99.8 & 5.6	& 0.2 & 11.8 & 0.2 & 0.0 & 5.3 & 1.1 & 2.2 & 2.5 & 6.4 & 11.2 & 17.4 & 12.6 & 61.3\\
\quad + H$_2$O & 4.2 & 4.2 & 4.2 & 4.2 & 0.0 & 0.0 & 0.8 & 0.4 & 0.3 & 1.2 & 7.1 & 8.6 & 17.4 & 4.0 & 61.3\\
\quad + Pyramid H$_2$O & 4.2 & 4.2 & 4.2 & 4.6 & 0.0 & 0.0 & 1.4 & 0.6 & 0.3 & 0.9 & 6.5 & 9.6 & 17.4 & 4.1 & 61.3\\
Llama3-G      & 99.4 & 18.0 & 21.0 & 13.0 & 0.0 & 0.2 & 5.3 & 2.1 & 1.6 & 3.2 & 12.7 & 9.6 & 14.6 & 15.4 & 613.3\\
\cellcolor{customblue}\textbf{UA, p=0} & \cellcolor{customblue} 100.0 & \cellcolor{customblue} 36.2 & \cellcolor{customblue} 36.4 & \cellcolor{customblue} 26.4 & \cellcolor{customblue} 6.4 & \cellcolor{customblue} 1.0 & \cellcolor{customblue} 16.1 & \cellcolor{customblue} 5.7 & \cellcolor{customblue} 30.5 & \cellcolor{customblue} 1.3 & \cellcolor{customblue} 0.0 & \cellcolor{customblue} 9.8 & \cellcolor{customblue} 12.2 & \cellcolor{customblue} 21.7 & \cellcolor{customblue}	 613.3\\
\cellcolor{customblue}\textbf{UA, p=0.05} & \cellcolor{customblue} 98.0 & \cellcolor{customblue} 25.8 & \cellcolor{customblue} 32.8  & \cellcolor{customblue} 23.4  & \cellcolor{customblue} 6.8  & \cellcolor{customblue} 1.0  & \cellcolor{customblue} 12.5  & \cellcolor{customblue} 4.9  & \cellcolor{customblue} 31.0  & \cellcolor{customblue} 0.2  & \cellcolor{customblue} 0.0  & \cellcolor{customblue} 8.2  & \cellcolor{customblue} 13.6  & \cellcolor{customblue} 19.9 & \cellcolor{customblue}	 24.0\\
\cellcolor{customblue}\textbf{UA, p=0.15} & \cellcolor{customblue} 87.4 & \cellcolor{customblue} 6.8 & \cellcolor{customblue} 13.0 & \cellcolor{customblue} 11.4 & \cellcolor{customblue} 3.6 & \cellcolor{customblue} 1.0 & \cellcolor{customblue} 8.5 & \cellcolor{customblue} 4.3 & \cellcolor{customblue} 35.0 & \cellcolor{customblue} 0.0 & \cellcolor{customblue} 0.0 & \cellcolor{customblue} 7.4 & \cellcolor{customblue} 12.2 & \cellcolor{customblue} 14.7 & \cellcolor{customblue} 13.5\\
\midrule\bottomrule
\end{tabular}
}
\end{table}

\end{document}